\documentclass[11pt]{article}

\usepackage[preprint]{acl}

\usepackage[utf8]{inputenc}
\usepackage[T1]{fontenc}
\usepackage{times}
\usepackage{amsmath,amssymb,amsthm}
\usepackage{mathtools}
\usepackage{booktabs}
\usepackage{multirow}
\usepackage{graphicx}
\usepackage{xcolor}
\usepackage{hyperref}
\usepackage{enumitem}
\usepackage{algorithm}
\usepackage{algpseudocode}
\usepackage{natbib}
\usepackage{microtype}
\usepackage{caption}
\usepackage{subcaption}
\usepackage{tcolorbox}
\tcbuselibrary{skins} 
\usepackage{xcolor}   

\setlist{nosep,leftmargin=1em}

\newcommand{\Halpha}{SRE}
\newcommand{\Hone}{SRE_{1}}
\newcommand{\Hord}{SRE_{\alpha}}
\newcommand{\Hhalf}{SRE_{1/2}}
\newcommand{\Zalpha}{\widehat{Z}_\alpha}
\newcommand{\ptheta}{p_\theta}
\newcommand{\Renyi}{R\'{e}nyi}
\newcommand{\eg}{\textit{e.g.}}

\newtheorem{proposition}{Proposition}
\newtheorem{lemma}{Lemma}

\newtcolorbox[auto counter]{ObservationBox}[2][]{%
  aibox,
  title={Observation~\thetcbcounter},
  label={#1},
  #2
}

\tcbset{
  aibox/.style={
    width=\linewidth,
    top=7pt,
    bottom=3pt,
    boxsep=1pt,
    left=2pt,
    right=2pt,
    colback=blue!6!white,
    colframe=black,
    colbacktitle=black,
    enhanced,
    center,
    attach boxed title to top left={yshift=-0.1in,xshift=0.1in},
    boxed title style={boxrule=0pt,colframe=white,},
  }
}
\newtcolorbox{AIbox}[2][]{aibox,title=#2,label=#1}

\title{Free Lunch Uncertainty: Sequence-Level \Renyi{} Entropy \\ from Particle Decoding for Adaptive LLM Reasoning}

\author{
Erfan Baghaei Potraghloo\textsuperscript{\rm 1}\thanks{Equal contribution},
Milad Heidari Sorkhehdizaj\textsuperscript{\rm 1}\footnotemark[1],
Shahab Rahimirad\textsuperscript{\rm 2}\footnotemark[1],
Aliasghar Pourghani\textsuperscript{\rm 2}\footnotemark[1],
\\
{\bf
 Seyedarmin Azizi\textsuperscript{\rm 1},
 Souvik Kundu \textsuperscript{\rm 3}
 , Abolfazl Hashemi \textsuperscript{\rm 2}, Joseph Campbell \textsuperscript{\rm 2}, Vijay Gupta \textsuperscript{\rm 2}
 }
 \\
\textsuperscript{\rm 1} University of Southern California,
\textsuperscript{\rm 2}Purdue University,
\textsuperscript{\rm 3}Intel
}
\date{}

\begin{document}
\maketitle
\raggedbottom

\begin{abstract}

Quantifying uncertainty over an entire LLM completion is critical for selective prediction and adaptive compute. Zero-cost signals such as mean token entropy and log-probability are local to tokens or to one completion, and self-consistency needs many extra completions. We show that Sequential Monte Carlo (SMC) decoding for power sampling already estimates the normalizing constant of the power distribution, which gives the sequence-level R\'enyi entropy (SRE) of the model's completion distribution at zero marginal cost, with a guarantee at every particle budget. A R\'enyi chain rule extends this readout to a family of stepwise entropies of any order. We prove that orders above one are tied to the most likely token of each step, which explains why they track log-probability, while the order-1/2 member accumulates the log of per-step search effort. This member is the best zero-cost selective-prediction signal in 10 of 13 settings. On MATH500 its AUROC is 0.022 to 0.044 above log-probability with all four models and within length quartiles, and the same order is selected when each setting is held out. An 8-particle probe identifies the problems that benefit from more particles, and a two-pass policy matches 64-particle accuracy on MATH500 with 34.2\% fewer particles over three seeds.
\end{abstract}

\section{Introduction}

Recent work suggests that reinforcement learning primarily surfaces capabilities already present in the base model's distribution, raising the probability of correct answers rather than teaching new reasoning skills \citep{yue2025does}.
This insight has motivated training-free alternatives that sharpen the model's distribution \citep{karan2025reasoning,azizi2026power,ji2026scalable}.
One such approach is power sampling: drawing completions $y$ for a prompt $x$ from a power distribution $\pi_\alpha(y|x) \propto \ptheta(y|x)^\alpha$, where $\ptheta$ is the language model and the sharpening exponent $\alpha > 1$ concentrates probability mass on higher-likelihood completions.

\begin{figure}[t]
    \centering
    \includegraphics[width=\linewidth]{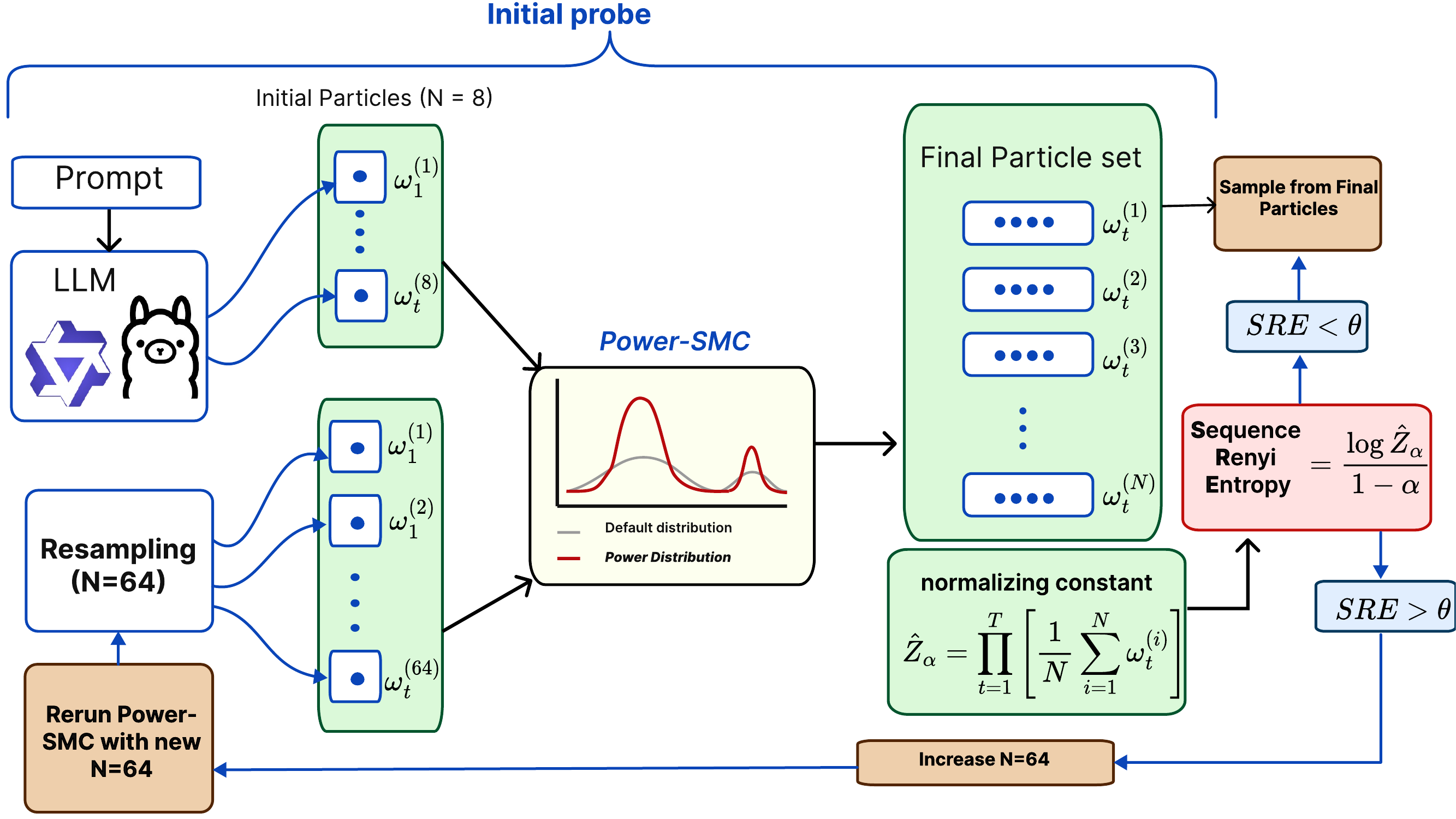}
   
    \caption{Overview of adaptive particle allocation using SRE.
SMC importance weights from an initial low-budget probe run are used to estimate the normalizing constant $Z_\alpha(x)$ and from it the sequence-level Rényi entropy.
High-SRE problems are re-run with a larger particle budget ($N=64$), while low-SRE problems keep the probe answer.}
\label{fig:overview}
\end{figure}

Power-SMC \citep{azizi2026power} and related methods \citep{nguyen2026apps} implement power sampling with Sequential Monte Carlo (SMC), maintaining $N$ parallel candidate completions (called \emph{particles} in the SMC literature) that are extended token by token, reweighted by importance weights, and resampled.
These methods perform well on reasoning tasks but use a fixed particle budget $N$ for every problem, regardless of difficulty.

A complementary challenge is \textbf{uncertainty quantification}: knowing when the model's answer should be trusted.

Uncertainty matters for selective prediction, calibration-aware deployment, adaptive compute allocation, and routing hard problems to stronger systems \citep{kadavath2022language}.
For reasoning tasks, uncertainty is naturally a property of the full completion distribution.
A model may be locally uncertain at a few tokens but follow one stable reasoning plan, or it may assign comparable probability to several incompatible solutions.

Existing uncertainty signals miss part of this structure.
Mean token entropy is zero-cost but measures local next-token ambiguity rather than uncertainty over whole completions \citep{kuhn2023semantic}.
Sequence log-probability scores only the returned completion and cannot distinguish a single dominant completion from one high-probability completion among several competing alternatives.
Self-consistency \citep{wang2023selfconsistency} and semantic entropy \citep{kuhn2023semantic,Farquhar2024DetectingHI} capture answer-level disagreement but require $K$ independent completions.

We make a simple observation: the SMC importance weights already give an estimate of the normalizing constant $Z_\alpha(x) = \sum_y \ptheta(y|x)^\alpha$, which defines the \textbf{sequence-level \Renyi{} entropy} of the completion distribution (\Renyi{} entropy generalizes Shannon entropy).
We call the resulting entropy estimate \textbf{SRE}.
When an SMC sampler is already running, this estimate gives a \textbf{zero-marginal-cost uncertainty signal} with no extra completions, classifiers, reward models, or model calls.

Beyond uncertainty quantification, $\Halpha$ enables \textbf{adaptive particle allocation}.
A cheap probe run with $N{=}8$ particles estimates $\Halpha$ per problem and predicts whether more particles will improve the answer.
The probe supplies both the signal and a candidate answer, so the allocation decision requires no additional model call. Figure~\ref{fig:overview} gives an overview.

\textbf{Our contributions.}
\begin{enumerate}

    \item We identify the SMC normalizing-constant estimate as a sequence-level \Renyi{} entropy signal for LLM reasoning, with an upward-bias and finite-$N$ guarantee that covers the sampler as run (Lemma~\ref{lem:smc_renyi}). The readout costs one scalar accumulation per decoding step, leaves the sampler unchanged, and applies to every SMC sampler whose final target is $\pi_\alpha$ (\S\ref{sec:method}).
    \item We use the \Renyi{} chain rule (Proposition~\ref{prop:chain}) to extend the readout to a family of stepwise entropies of any order and characterize the family (Proposition~\ref{prop:spectrum}). Each member accumulates the negative log of a power mean of the probability that the model gives to its own next token; members of order $\beta$ above one are tied to the top token within a factor $\beta/(\beta-1)$; and the order-1/2 member accumulates the log of the per-step search effort of Proposition~\ref{prop:search}.
    \item We show that the order-1/2 member alone is the best zero-cost selective-prediction signal in 10 of 13 settings spanning four models, two benchmarks, and six sharpening exponents (\S\ref{sec:selective}, \S\ref{sec:alpha_ablation}). Its area under the ROC curve (AUROC) is 0.022 to 0.044 above that of log-probability on MATH500 with all four models ($p \leq 0.029$ after Holm correction). The margin holds within length quartiles and with the order selected on held-out settings, and the order-1/2 member is also above a generation-length baseline (\S\ref{sec:order}, \S\ref{sec:analysis}).
    \item We show that $\Halpha$ from a cheap $N{=}8$ probe predicts the gain from additional particles (\S\ref{sec:killswitch}). Across three seeds and two models, the low-$\Halpha$ tercile gains at most 0.6 points from $N{=}64$ while the upper two terciles together gain 1.2 to 4.8 points, a pattern that an independent $N{=}8$ re-draw does not reproduce. With Qwen2.5-Math-7B a two-pass policy matches uniform $N{=}64$ accuracy on MATH500 with 34.2\% fewer particles on average over three seeds (\S\ref{sec:pareto}).
\end{enumerate}

\section{Background}
\label{sec:back}

\subsection{Power-SMC}

Given a prompt $x$, an autoregressive language model defines $\ptheta(y|x) = \prod_{t=1}^{T} \ptheta(y_t \mid x,y_{<t})$ over completions $y$ of length $T$ that end with the end-of-sequence (EOS) token.
Power sampling draws completions from the \emph{power distribution}
\begin{equation}
    \pi_\alpha(y|x)
    =
    \frac{\ptheta(y|x)^\alpha}{Z_\alpha(x)},
    \quad
    Z_\alpha(x)=\sum_y \ptheta(y|x)^\alpha,
\end{equation}
where $\alpha > 1$ is the sharpening exponent.

Power-SMC \citep{azizi2026power} maintains $N$ particles and samples each token from a \emph{proposal distribution}: the locally optimal choice $q_t^\star(v) \propto \ptheta(v|x,y_{<t}^{(i)})^\alpha$, which corresponds to sampling at temperature $\tau{=}1/\alpha$ and minimizes the variance of the incremental weight below.
Under this proposal, the incremental importance weight (incremental weight for short) for particle~$i$ at step $t$ is
\begin{equation}
\label{eq:omega}
    \omega_t^{(i)}
    =
    \sum_v
    \ptheta(v|x,y_{<t}^{(i)})^\alpha,
\end{equation}
which is deterministic given the prefix \citep{azizi2026power}.
When the effective sample size (ESS, a measure of how many particles carry non-negligible weight) drops below a threshold, Power-SMC performs systematic resampling: duplicating high-weight particles and discarding low-weight ones to prevent weight collapse onto a few particles.

\subsection{\Renyi{} entropy}

The \Renyi{} entropy of order $\alpha$ of the full completion distribution $\ptheta(\cdot|x)$ is
\begin{equation}
\label{eq:renyi}
    H_\alpha\!\left(\ptheta(\cdot|x)\right)
    =
    \frac{1}{1-\alpha}
    \log
    \underbrace{
    \sum_y \ptheta(y|x)^\alpha
    }_{Z_\alpha(x)}.
\end{equation}
For $\alpha > 1$, $H_\alpha$ is low when probability mass concentrates on a few completions and high when mass spreads across many.
As $\alpha \to 1$, $H_\alpha$ converges to the Shannon entropy.

\section{Motivating Analysis}
\label{sec:motive}

\subsection{Analysis 1: Sequence-level entropy is a sum, not a mean}
\label{sec:analysis1}

Mean token entropy, $\frac{1}{T}\sum_t H(Y_t\mid Y_{<t},x)$ along a generated completion, is a common zero-cost signal of predictive uncertainty in language models \citep{malinin2021uncertainty}.
The entropy of the completion distribution is not an average of next-token entropies.
For Shannon entropy the chain rule gives $H(Y_{1:T}\mid x) = \sum_{t=1}^{T} \mathbb{E}_{y_{<t}\sim\ptheta}\left[H(Y_t\mid y_{<t},x)\right]$, a sum over steps \citep{cover2006elements,malinin2021uncertainty}.
\Renyi{} entropy obeys an analogous rule \citep{jizba2004world}.
Let $\pi_\beta^{(t)}(y_{1:t}) \propto \ptheta(y_{1:t}|x)^\beta$ be the order-$\beta$ power distribution over length-$t$ prefixes and $h_\beta(y_{<t}) = \frac{1}{1-\beta}\log\sum_v \ptheta(v|x,y_{<t})^\beta$ the \Renyi{} entropy of the next-token distribution.

\begin{proposition}[\Renyi{} chain rule]
\label{prop:chain}
For $\beta > 0$, $\beta \neq 1$, and prefixes $Y_{<t}\sim\pi_\beta^{(t-1)}$,
\begin{equation*}
H_\beta\!\left(\ptheta\right) = \sum_{t=1}^{T} \frac{1}{1-\beta} \log \mathbb{E}\!\left[e^{(1-\beta)\,h_\beta(Y_{<t})}\right],
\end{equation*}
where $H_\beta(\ptheta)$ abbreviates $H_\beta(\ptheta(\cdot|x))$, and the right-hand side converges to the right-hand side of the Shannon chain rule as $\beta \to 1$.
\end{proposition}

The proof is in Appendix~\ref{app:chain}.
Sequence-level entropy is therefore a sum of per-step increments, each an aggregate of next-token entropies over prefixes.
Dividing by the length, as mean token entropy does, removes the accumulation that separates a long uncertain completion from a short one, and following a single completion ignores the expectation over prefixes.

\subsection{Analysis 2: SMC importance weights already contain distributional information}
\label{sec:analysis2}

\textbf{Analysis.}
Classical SMC theory provides an unbiased estimator $\Zalpha$ of the normalizing constant, $\mathbb{E}[\Zalpha]=Z_\alpha$ \citep{delmoral2004feynman,chopin2020introduction}.
We call it the standard estimator:
\begin{equation}
\label{eq:zhat}
    \log \Zalpha
    =
    \sum_{t=1}^{T}
    \log
    \left[
        \sum_{i=1}^{N}
        W_{t-1}^{(i)}\,
        \omega_t^{(i)}
    \right],
\end{equation}
where $W_{t-1}^{(i)}$ are the normalized weights before step $t$, equal to $1/N$ at the start and after every resampling step.
Power-SMC already computes every $\omega_t^{(i)}$, which follows Eq.~\ref{eq:omega} once an initial ramp of the exponent has ended, and every $W_t^{(i)}$ as part of its usual importance-weight computation.
The SMC sampler therefore \emph{already computes an estimate of $Z_\alpha$}, one step at a time, but it discards this information.

\section{Methodology}
\label{sec:method}

\subsection{Extracting \texorpdfstring{SRE}{SRE} from Power-SMC}

Because the logarithm is nonlinear, $\log\Zalpha$ is a plug-in estimate of $\log Z_\alpha$ rather than an unbiased estimator of it.

We define the estimate of the \textbf{sequence-level \Renyi{} entropy}, \textbf{SRE}, as
\begin{equation}
    \Halpha
    =
    \frac{\log \Zalpha}{1-\alpha}.
\end{equation}

Computing $\Zalpha$ requires only accumulating one scalar per decoding step: the log of the weighted mean of the incremental weights across particles.

\textbf{Stepwise accumulation.}
In all experiments we accumulate the log of the uniform mean of the incremental weights over particles at every step,
\begin{equation}
\label{eq:zstep}
    \log \Zalpha^{\mathrm{step}}
    =
    \sum_{t=1}^{T}
    \log
    \left[
        \frac{1}{N}
        \sum_{i=1}^{N}
        \omega_t^{(i)}
    \right],
\end{equation}
which is Eq.~\ref{eq:zhat} with $W_{t-1}^{(i)}{=}1/N$.
It coincides with Eq.~\ref{eq:zhat} when resampling occurs at every step.
Under the blockwise resampling of Power-SMC (resampling at block boundaries when the ESS is low), the stepwise accumulation uses the next-token distributions of all $N$ particles at every step, rather than mainly those of the highest-weight particles.
The two estimates have a Spearman rank correlation of at least 0.98 in every setting we test, and Appendix~\ref{app:variants} compares them.
Algorithm~\ref{alg:extract} in Appendix~\ref{app:resampling} gives the implementation and the handling of resampling boundaries.

\textbf{Stepwise family.}
Once the exponent ramp of Power-SMC has ended (Appendix~\ref{app:resampling}), $\omega_t^{(i)} = e^{(1-\alpha)h_\alpha(y_{<t}^{(i)})}$ up to the nucleus truncation of the proposal, which drops the least likely tokens.
Each term of Eq.~\ref{eq:zhat} is therefore, up to a constant factor, the particle estimate of one increment of the order-$\alpha$ chain rule in Proposition~\ref{prop:chain}.
The weighted particles follow the intermediate target of the sampler, and Eq.~\ref{eq:zstep} replaces the weighted mean by the uniform mean.
The same per-step accumulation can be computed at any order $\beta$ from the next-token distributions that the sampler has already computed:
\begin{equation}
\label{eq:family}
    SRE_{\beta}
    =
    \sum_{t=1}^{T}
    \tfrac{1}{1-\beta}
    \log\!\Big[
        \tfrac{1}{|A_t|}
        \sum_{i\in A_t}
        e^{(1-\beta)h_\beta(y_{<t}^{(i)})}
    \Big],
\end{equation}
where $A_t$ is the set of particles still generating at step $t$, with limit $\Hone = \sum_t \frac{1}{|A_t|}\sum_{i\in A_t} H(\ptheta(\cdot|x,y_{<t}^{(i)}))$ as $\beta \to 1$.

Whatever the order, the particles are drawn by the $\alpha$-sharpened sampler, so $SRE_{\beta}$ measures order-$\beta$ entropy accumulated along the particles.

At $\beta{=}\alpha$ this family member uses no ramp term and agrees with the stepwise accumulation of Eq.~\ref{eq:zstep} at Spearman $\rho \geq 0.94$ (Appendix~\ref{app:variants}).

\textbf{Choice of order.}
$SRE_\beta$ is the accumulated negative log of a power mean, of order $\beta-1$, of the probability that the model gives to a token drawn from its own next-token distribution (Proposition~\ref{prop:spectrum}, Appendix~\ref{app:spectrum}).
It is nonincreasing in $\beta$ and runs from a multiple of the length ($\beta\to0$) to the summed negative log-probability, at each step, of the most likely top token among the particles ($\beta\to\infty$).
For every $\beta>1$ each increment lies within a factor $\beta/(\beta-1)$ of the top-token term, so members of these orders are tied to the top token; members of order up to one also respond to how the remaining probability mass is spread.
At $\beta{=}1$ the family gives $\Hone$, the chain-rule sum of Shannon entropies \citep{malinin2021uncertainty} on the particle set, and the order-$1/2$ member accumulates the log of the per-step search effort defined next.

\begin{proposition}[Search effort]
\label{prop:search}
Let $Y\sim p$ be the completion to be found, and let a sampler draw completions independently from a proposal $q$.
The expected number of draws until $Y$ is first drawn is at least $e^{H_{1/2}(p)}$ for every $q$, with equality for $q \propto p^{1/2}$.
\end{proposition}

The proof is in Appendix~\ref{app:search}; the same quantity governs guessing in order of decreasing probability \citep{arikan1996inequality}.
Applied to the next-token distribution, $e^{h_{1/2}(y_{<t})}$ is the least expected number of independent draws needed to hit the model's next token, and $\Hhalf$ accumulates the log of this per-step search effort along the particles.
The stepwise accumulation of order $\alpha$ has an analogous reading for the power distribution: $H_\alpha(\ptheta)=H_{1/\alpha}(\pi_\alpha)$, and $e^{H_\alpha(\ptheta)}$ bounds the order-$(\alpha{-}1)$ guesswork of a completion drawn from $\pi_\alpha$ within a factor $1+\ln M$, where $M$ is the number of possible completions (Appendix~\ref{app:search}).
We report three SRE signals: $\Halpha$, the stepwise accumulation of order $\alpha$ (Eq.~\ref{eq:zstep}), which comes from the incremental weights with no further computation; $\Hone$; and $\Hhalf$.

\subsection{Theoretical guarantee}

\begin{lemma}[SMC estimator of $Z_\alpha$]
\label{lem:smc_renyi}
Consider an SMC sampler with $N$ particles whose final target is $\pi_\alpha$, with any intermediate targets and proposals and with unbiased resampling at times that may depend on the current particle system.
Let $\widehat{Z}_{\alpha,N}$ be the standard estimator of Eq.~\ref{eq:zhat} and $\widehat{\mathrm{SRE}}_{\alpha,N} = \frac{1}{1-\alpha}\log\widehat{Z}_{\alpha,N}$.
For every $N \geq 1$, $\alpha > 1$, and $\delta \in (0,1)$:
\begin{enumerate}[leftmargin=1.5em]
    \item $\mathbb{E}[\widehat{Z}_{\alpha,N}] \leq Z_\alpha$, with equality for full-support proposals;
    \item $\mathbb{E}[\widehat{\mathrm{SRE}}_{\alpha,N}] \geq H_\alpha(\ptheta(\cdot|x))$; \hfill (upward bias)
    \item $\Pr\big[\widehat{\mathrm{SRE}}_{\alpha,N} \geq H_\alpha(\ptheta(\cdot|x)) - \tfrac{\log(1/\delta)}{\alpha-1}\big] \geq 1-\delta$. \hfill (finite $N$)
\end{enumerate}
With full-support proposals and multinomial resampling at fixed times, $\widehat{\mathrm{SRE}}_{\alpha,N} \to H_\alpha(\ptheta(\cdot|x))$ in probability as $N\to\infty$.
\end{lemma}

The proof is in Appendix~\ref{app:proof}.
The bounds of the lemma cover Power-SMC as run, with its exponent ramp, its nucleus-truncated proposal, and its ESS-triggered resampling.
The guarantee is non-asymptotic: at $\alpha{=}4$ the estimate is more than one nat below the true sequence-level \Renyi{} entropy in at most 5\% of per-problem runs, for every particle budget including $N{=}8$.

\textbf{Generality.} Lemma~\ref{lem:smc_renyi} holds for every SMC sampler whose final target is $\pi_\alpha$, whatever its positive intermediate targets and its unbiased resampling rule. The readout therefore applies, at the same zero marginal cost, to other SMC samplers for power sampling such as APPS \citep{nguyen2026apps} (Appendix~\ref{app:generality}).

\subsection{Adaptive particle allocation}
\label{sec:adaptive_method}

Fixed particle budgets waste compute: easy problems are solved with a small $N$, while hard problems may need many particles.

We use a two-pass policy with a threshold $\theta$: run Power-SMC with $N_{\text{probe}}$ particles (\eg, $N{=}8$) and extract $\Halpha$ and the candidate answer; if $\Halpha > \theta$, escalate the problem by re-running it with $N_{\text{full}}$ (\eg, $N{=}64$), and otherwise accept the probe answer.
The probe supplies the signal before committing to the expensive run, and its own answer serves as the fallback.

\section{Experiments}

\subsection{Setup}

We evaluate on \textbf{MATH500} \citep{hendrycks2021measuring,lightman2023verify} and \textbf{GSM8K} \citep{cobbe2021training} for mathematical reasoning.
We test four models from two model families:
\textbf{Qwen2.5-Math-1.5B}, \textbf{Qwen2.5-Math-7B} (math-specialized), \textbf{Qwen2.5-7B} (general-purpose) \citep{qwen2.5}, and
\textbf{Llama-3.1-8B-Instruct} \citep{dubey2024llama}.
For each problem, we extract six uncertainty signals at zero marginal cost from the SMC run:
(1)~$\Hhalf$ and (2)~$\Hone$ from Eq.~\ref{eq:family} (ours);
(3)~$\Hord$ from Eq.~\ref{eq:zstep} (ours);
(4)~mean token entropy, which is $\Hone$ divided by the number of decoding steps;
(5)~sequence log-probability, $\log\ptheta(y|x)$, of the returned completion;
(6)~particle agreement, the answer agreement across the $N$ SMC particles.\footnote{SMC particles share ancestors after resampling, so this is correlated self-consistency (Appendix~\ref{app:sc}).}

None of them requires decoding beyond the Power-SMC run; generation length is reported as a reference. 
The SMC runs use Hugging Face Transformers with bfloat16 model parameters on NVIDIA RTX A6000 GPUs. The base performance of Power-SMC is reported in Appendix~\ref{app:smcbaseline}. We will release the readout patch, the evaluation scripts, and per-problem logs.
\subsection{Selective prediction}
\label{sec:selective}

\textbf{Experiment.}
We rank problems by each signal and report the AUROC for predicting the correctness of the returned answer in eight model--benchmark settings (Table~\ref{tab:main}); lower SRE, lower mean token entropy, higher log-probability, and higher answer agreement count as higher confidence.

\begin{table*}[t]
\centering
\small
\begin{tabular}{@{}lcccccccc@{}}
\toprule
 & \multicolumn{4}{c}{\textbf{MATH500}} & \multicolumn{4}{c}{\textbf{GSM8K}} \\
\cmidrule(lr){2-5} \cmidrule(lr){6-9}
\textbf{Signal} & \textbf{M-1.5B} & \textbf{M-7B} & \textbf{Q-7B} & \textbf{Llama} & \textbf{M-1.5B} & \textbf{M-7B} & \textbf{Q-7B} & \textbf{Llama} \\
\midrule
$\Hhalf$ (ours) & \textbf{.836} & \textbf{.850} & \textbf{.834} & \textbf{.884} & \textbf{.776} & \underline{.782} & .658 & \textbf{.743} \\
$\Hone$ (ours) & \underline{.825} & \underline{.821} & \underline{.824} & \underline{.866} & \underline{.760} & .779 & .686 & \underline{.733} \\
$\Hord$ (ours) & .816 & .806 & .809 & \underline{.866} & .748 & .753 & .642 & .724 \\
Seq log-prob & .810 & .805 & .801 & .862 & .744 & .741 & .635 & .719 \\
Self-cons.$^\dag$ & .656 & .696 & .650 & .547 & .718 & .713 & \underline{.688} & .626 \\
Mean token entropy & .714 & .699 & .531 & .390 & .728 & \textbf{.820} & \textbf{.697} & .691 \\
Generation length & .690 & .678 & .722 & .852 & .592 & .479 & .493 & .626 \\
\bottomrule
\multicolumn{9}{@{}l}{\scriptsize $^\dag$Correlated SMC-particle agreement (Appendix~\ref{app:sc}).}
\end{tabular}

\caption{Selective prediction AUROC ($N{=}32$, $\alpha{=}4$, 500 problems per setting). M: Qwen2.5-Math, Q-7B: Qwen2.5-7B, Llama: Llama-3.1-8B-Instruct. $\Hord$: stepwise accumulation of order $\alpha$ from the incremental weights (Eq.~\ref{eq:zstep}); $\Hone$ and $\Hhalf$: order-1 and order-1/2 members (Eq.~\ref{eq:family}). Best in \textbf{bold}, second best \underline{underlined}. Bootstrap confidence intervals in Appendix~\ref{app:bootstrap}.}
\label{tab:main}
\end{table*}

$\Hhalf$ is the best zero-cost signal in six of the eight settings.
On MATH500, the SRE signals are the best for all four models, with $\Hhalf$ between 0.834 and 0.884.
With the larger Qwen2.5-14B-Instruct the SRE signals reach 0.863 to 0.867, against 0.860 for log-probability, and with the newer Qwen3-4B \citep{yang2025qwen3}, $\Hhalf$ reaches 0.864 against 0.837 ($p < 0.001$; Appendix~\ref{app:larger}).

On GSM8K, where generation length carries little information about correctness (AUROC 0.48 to 0.63, against 0.68 to 0.85 on MATH500), length normalization matters less: mean token entropy leads on Qwen2.5-Math-7B and Qwen2.5-7B, by margins over $\Hhalf$ that are not significant ($p{=}0.19$ and $0.33$).
The order-1/2 member is the best signal on Qwen2.5-Math-1.5B and Llama-3.1-8B and the second best on Qwen2.5-Math-7B.

$\Hone$ is above sequence log-probability in all eight settings, by 0.004 to 0.051 AUROC, and the difference is significant in six of them (paired permutation tests, $p < 0.05$; Appendix~\ref{app:bootstrap}). $\Hord$, which needs no computation beyond the incremental weights, is above log-probability in all eight as well, by up to 0.013.
$\Hhalf$ is the strongest SRE signal in seven of the eight settings: on MATH500 it is 0.022 to 0.044 above log-probability ($p \leq 0.006$; $p \leq 0.029$ after Holm correction over the eight settings), and on GSM8K it is 0.023 to 0.041 above (\S\ref{sec:order}).

\begin{figure}[t]
  \centering
  \includegraphics[width=\columnwidth]{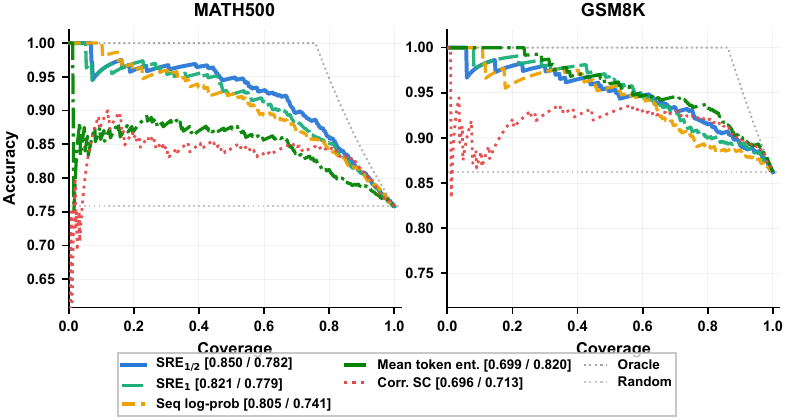}

  \caption{Selective prediction with Qwen2.5-Math-7B: accuracy vs.\ coverage on MATH500 (\textbf{left}) and GSM8K (\textbf{right}). AUROC in legend as [MATH500\,/\,GSM8K].}
  \label{fig:acc_cov}
  \vspace{-4mm}
\end{figure}

Abstaining on the 20\% highest-$\Hhalf$ problems raises accuracy on the rest from 75.8\% to 86.0\% (MATH500, Qwen2.5-Math-7B; Figure~\ref{fig:acc_cov}).

Combining $\Hhalf$ with the three baselines in a logistic regression gives 0.878 cross-validated AUROC (Appendix~\ref{app:combination}).

\subsection{Does SRE predict the gain from more particles?}
\label{sec:killswitch}

\textbf{Experiment.}
We run Power-SMC at $N \in \{8, 32, 64\}$ on MATH500 with Qwen2.5-Math-7B and partition the problems into terciles of $\Hord$ from the $N{=}8$ run (Table~\ref{tab:killswitch}).

\begin{table}[t]
\centering
\small
\begin{tabular}{@{}lrcccr@{}}
\toprule
\textbf{Tercile} & $n$ & $N{=}8$ & $N{=}32$ & $N{=}64$ & \textbf{Gain} \\
\midrule
Low & 167 & 94.0 & 93.4 & 92.8 & $-$1.2 \\
Mid & 166 & 83.7 & 84.3 & 86.7 & $+$3.0 \\
High & 167 & 46.7 & 49.7 & 52.1 & $+$5.4 \\
\midrule
\textit{All} & \textit{500} & \textit{74.8} & \textit{75.8} & \textit{77.2} & $+$\textit{2.4} \\
\bottomrule
\end{tabular}
\caption{Accuracy (\%) by $\Hord$ tercile from the $N{=}8$ probe (500 problems, Qwen2.5-Math-7B, $\alpha{=}4$). \textbf{Gain} = $\mathrm{Acc}(N{=}64) - \mathrm{Acc}(N{=}8)$.}
\label{tab:killswitch}
\end{table}

The gain from more particles grows with probe $\Hord$: $-$1.2, $+$3.0, and $+$5.4 points from the low to the high tercile.

\paragraph{Replication.}
Averaged over three seeds, the gains are $-$0.6, $+$2.4, and $+$2.6 points, and on Qwen2.5-Math-1.5B they are $+$0.6, $+$3.6, and $+$6.0.
In every run the low tercile gains at most 0.6 points, while the upper two terciles together gain 1.2 to 4.8 points (Appendix~\ref{app:sweep}).
This rise across terciles is not a re-draw effect: over all pairings of the three seeds, an independent $N{=}8$ run changes accuracy by $-$0.3, $+$0.4, and $-$0.1 points in the three terciles, against $-$0.5, $+$2.1, and $+$2.8 for $N{=}64$, and in the upper two terciles $N{=}64$ gains 2.3 points more than a re-draw (95\% confidence interval 0.2 to 4.4).

\paragraph{Rank stability.}
The Spearman correlation between $\Hord$ estimated at $N{=}8$ and $N{=}64$ is 0.953 to 0.965 on the three seeds, and probe $\Hhalf$ identifies the problems that are correct at both particle budgets with AUROC 0.841 to 0.845, against 0.808 to 0.818 for log-probability (Appendix~\ref{app:transitions}).

\subsection{Which entropy order?}
\label{sec:order}

\textbf{Experiment.}

\begin{figure}[t]
  \centering
  \includegraphics[width=\columnwidth]{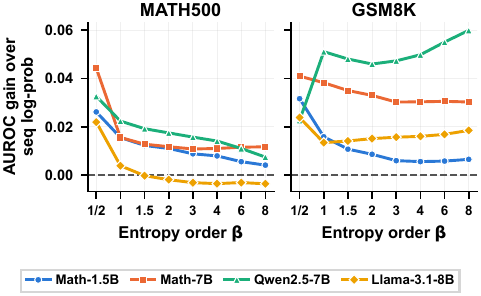}

  \caption{Selective prediction AUROC of $SRE_\beta$ (Eq.~\ref{eq:family}) minus that of sequence log-probability, as a function of the order $\beta$ ($N{=}32$, $\alpha{=}4$).}
  \label{fig:order}
  \vspace{-4mm}
\end{figure}

We log $SRE_\beta$ for orders from $1/4$ to $\infty$ in the same runs and compare them on the same problems (Figure~\ref{fig:order}).

On MATH500 the margin over sequence log-probability is largest at order $1/2$ for every model: $\Hhalf$ reaches 0.836, 0.850, 0.834, and 0.884 with the four models, against 0.825, 0.821, 0.824, and 0.866 for $\Hone$.
On GSM8K the order-1/2 member is the best of the family with three of the four models (0.776, 0.782, and 0.743, against 0.760, 0.779, and 0.733); with Qwen2.5-7B, $\Hone$ is higher (0.686 against 0.658).
At $\beta{=}\infty$, where only the top token of each step counts, AUROC falls below that of log-probability (0.615 to 0.769 on MATH500).
Below $1/2$ the tail of the vocabulary dominates every increment and $SRE_\beta$ approaches a multiple of the completion length (Spearman $\rho \geq 0.93$ at $\beta{=}1/4$), so AUROC drops, to 0.743 on MATH500 with Qwen2.5-7B.

The choice of order is not tuned to a setting: selecting the order by mean AUROC on the other seven settings picks $1/2$ for each held-out setting, and so do the four $N{=}8$ probe runs and the $N{=}64$ run of \S\ref{sec:pareto} that log every order of the stepwise family (0.845, 0.824, and 0.827 for $\Hhalf$ on the three probe seeds, against 0.813, 0.792, and 0.792 for log-probability).
$\Hhalf$ does not depend on the far tail of the vocabulary: computed over the 100 most likely tokens of each step, it is above log-probability in all seven settings where we log it, by 0.014 to 0.052, and over the 10 most likely tokens by 0.008 to 0.041.

\subsection{Sensitivity to \texorpdfstring{$\alpha$}{alpha}}
\label{sec:alpha_ablation}

We sweep $\alpha \in \{1.5, 2, 3, 4, 6, 8\}$ on MATH500 with Qwen2.5-Math-7B (Figure~\ref{fig:alpha_ablation}, Appendix~\ref{app:alpha}).
$\Hhalf$ is the best of the logged orders at every $\alpha$ and the best of all signals from $\alpha{=}2$ upward, 0.044 to 0.069 AUROC above sequence log-probability ($p \leq 0.001$); at $\alpha{=}1.5$, where particle agreement leads (0.831), it is 0.079 above.
$\Hone$ stays between 0.802 and 0.822 and is above log-probability at every $\alpha$ (all $p \leq 0.002$), while particle agreement falls to 0.598 at $\alpha{=}8$.

\subsection{Adaptive particle allocation: accuracy vs.\ compute}
\label{sec:pareto}

\textbf{Experiment.}
We apply the two-pass policy of \S\ref{sec:adaptive_method} to MATH500 with Qwen2.5-Math-7B: the probe runs at $N{=}8$ and problems with $\Hord$ above a threshold are re-run at $N{=}64$ (every escalation rate in Appendix~\ref{app:sweep}).

\begin{figure}[t]
  \centering
  \includegraphics[width=\columnwidth]{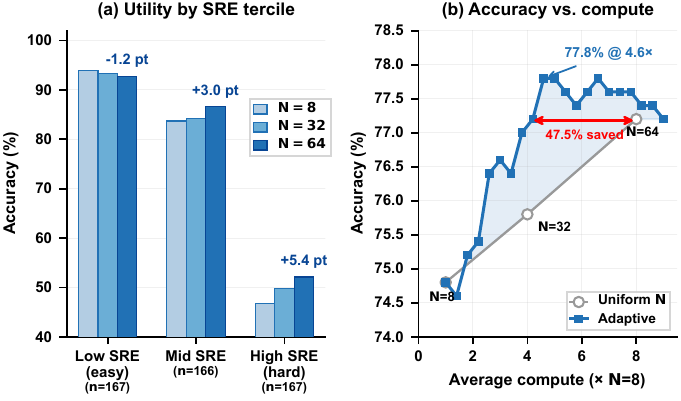}
  \caption{Adaptive particle allocation on MATH500 (Qwen2.5-Math-7B, $\alpha{=}4$).
  \textbf{(a)} Accuracy gain from $N{=}8{\to}64$ by $\Hord$ tercile.
  \textbf{(b)} Accuracy vs.\ compute: the two-pass policy reaches uniform $N{=}64$ accuracy from $4.2\times$ onward and peaks at 77.8\%.}
  \label{fig:pareto}
  \vspace{-4mm}
\end{figure}

Escalating the 40\% highest-$\Hord$ problems matches uniform $N{=}64$ accuracy (77.2\%) with 47.5\% fewer particles (Figure~\ref{fig:pareto}); on two further seeds the saving is 22.5\% and 32.5\%, for an average of 34.2\%.
With one escalation rate for all seeds, escalating 60\% gives the three-seed mean accuracy of uniform $N{=}64$ (76.6\%) with 27.5\% fewer particles, and random escalation saves nothing on average ($-$0.8\%; Appendix~\ref{app:sweep}).
With abstention added, a three-way policy (abstain, escalate, or keep the probe answer) matches uniform $N{=}64$ on the 80\% most confident problems (85.1\% vs.\ 85.0\%) with $4.2\times$ instead of $8\times$ the probe's particle budget (Appendix~\ref{app:threeway}).

\paragraph{Comparison of gating signals.}
Table~\ref{tab:gating} (Appendix~\ref{app:sweep}) compares $\Hord$ as a gating signal with other signals from the same probe run, including answer entropy over the probe's particles (used by Certaindex \citep{fu2025certaindex}) and Self-Certainty \citep{kang2025selfcertainty}.
Averaged over escalation rates and three seeds, every signal beats random escalation.
$\Hhalf$ is the best gating signal on the first seed, and $\Hord$ is within 0.94 points of the best gating signal in every run.

\subsection{Analysis}
\label{sec:analysis}

\paragraph{Why the order matters.}
For every order above one, each increment lies between the negative log-probability of the top token and $\beta/(\beta-1)$ times it, and at the sequence level, with $H_\infty$ the negative log-probability of the most likely completion, $H_\infty \le H_\alpha \le \frac{\alpha}{\alpha-1}H_\infty$ (Proposition~\ref{prop:spectrum}).
Members of such orders are tied to the top token, which is also what the log-probability of a completion sampled at low temperature reflects.
The data agree: at $\alpha{=}4$ the standard estimator is within 0.004 AUROC of log-probability in all eight settings, and the order-4 increments sum to 1.25 times the top-token terms (median, inside the range $[1, 4/3]$ of the proposition).
Members of order up to one also respond to how the remaining probability mass is spread, and this is where the margin appears: 0.004 to 0.051 at order 1 and 0.022 to 0.044 at order $1/2$.

\paragraph{The particle set.}
At order $1/2$, aggregating over the particle set raises AUROC above that of the same sum along the returned completion at every $\alpha$ with Qwen2.5-Math-7B on MATH500: by 0.062, 0.030, 0.030, 0.012, 0.012, and 0.011 for $\alpha{=}1.5$, 2, 3, 4, 6, and 8 ($p < 0.05$ up to $\alpha{=}3$).
This advantage grows with the diversity of the particles (2.1, 1.6, 1.4, 1.3, 1.2, and 1.2 distinct answers per problem); at $\alpha{=}4$ it is 0.020 with Qwen2.5-7B and 0.014 with Qwen3-4B ($p \leq 0.020$).
At equal log-probability, higher $\Hhalf$ goes with more disagreement among the particles' answers in all eight settings of Table~\ref{tab:main} (partial Spearman correlation 0.06 to 0.34).

\paragraph{Sum, not mean.}
$\Hone$ and mean token entropy are built from the same next-token entropies and differ only in the division by the length.
On MATH500 that normalization costs 0.11 to 0.48 AUROC: dividing by the length removes the accumulated part.
The accumulated part is more than length. Generation length alone reaches 0.68 to 0.85 AUROC on MATH500 and 0.48 to 0.63 on GSM8K, and $\Hhalf$ is 0.03 to 0.30 above it in all eight settings ($p \leq 0.001$ after Holm correction).
The same normalization also hurts at order $1/2$.
Dividing $\Hhalf$ by the length gives a particle-set version of the average per-token \Renyi{} score that \citet{darrin2023rainproof} use for out-of-distribution detection, and it lowers the AUROC from 0.850 to 0.708 (MATH500, Qwen2.5-Math-7B).

\paragraph{SRE is not a length proxy.}
Within every quartile of generation length, the half of problems with above-median $\Hord$ is less accurate, by 18 to 39 points on MATH500 and 13 to 29 points on GSM8K (Table~\ref{tab:length_strat}, Appendix~\ref{app:length}).
Inside length quartiles $\Hhalf$ stays above sequence log-probability in all eight settings, by 0.030 to 0.050 AUROC on MATH500 and 0.020 to 0.052 on GSM8K (Table~\ref{tab:length_strat_all}); the pooled margins are 0.022 to 0.044 and 0.023 to 0.041, so the advantage does not come from length.

\section{Related Work}

\textbf{LLM uncertainty quantification.}
Self-consistency \citep{wang2023selfconsistency} measures answer agreement across $K$ independent completions.
Semantic entropy \citep{kuhn2023semantic} clusters sampled completions by meaning.
P(True) \citep{kadavath2022language} asks the model in an extra query whether its answer is true.
\citet{malinin2021uncertainty} estimate sequence-level Shannon entropy from independent completions with the chain rule; $\Hone$ applies this chain-rule sum to the particle set of an SMC sampler, and Proposition~\ref{prop:chain} extends it to \Renyi{} orders.
\citet{darrin2023rainproof} average a per-token \Renyi{} divergence from the uniform distribution along one completion; SRE accumulates instead of averaging and aggregates over the particle set.
To our knowledge, $\Hord$ is the first sequence-level entropy signal for LLM generation built from the normalizing-constant estimate of an importance-weighted sampler.
Appendix~\ref{app:positioning} gives a detailed comparison.

\textbf{Power sampling for LLM reasoning.}
Power sampling was introduced with a Metropolis--Hastings sampler \citep{karan2025reasoning} and made faster with SMC \citep{azizi2026power}, rollouts (sampled continuations) \citep{ji2026scalable}, and future-value potentials \citep{nguyen2026apps}.
None of these works use the normalizing-constant estimate as an uncertainty signal or for adaptive particle allocation.

\textbf{Adaptive inference-time compute.}
Several works allocate compute based on problem difficulty \citep{snell2024scaling}, typically with trained difficulty predictors or reward models.
Certaindex \citep{fu2025certaindex} tracks the stability of intermediate answers, such as answer entropy across sampled completions, to stop or reallocate compute.
CaTS \citep{huang2026cats} distills self-consistency confidence into the model and uses it for early stopping in repeated sampling.
Self-Certainty \citep{kang2025selfcertainty} scores each completion by the divergence of its next-token distributions from uniform for best-of-$N$ selection.
SRE is complementary: it is read from the SMC sampler itself and gates the particle budget of that same sampler.
\section{Conclusions}

We showed that SMC decoding already estimates the sequence-level \Renyi{} entropy of the model's completion distribution at the cost of one scalar accumulation per decoding step, with a guarantee that holds at every particle budget.
A \Renyi{} chain rule turns this readout into a family of stepwise entropies whose order-1/2 member, which accumulates the log of per-step search effort, is the best zero-cost selective-prediction signal in 10 of 13 settings, with a margin over log-probability that holds within length quartiles and with the order chosen on held-out settings.
A cheap $N{=}8$ probe identifies which problems need expensive decoding, and the two-pass policy matches uniform $N{=}64$ accuracy on MATH500 with 34.2\% fewer particles on average over three seeds.

\section*{Limitations}

SRE is computed from the particle set of an SMC run and has zero marginal cost relative to that run (Appendix~\ref{app:sc} reports the order-1/2 sum along the returned completion alone); decoding that draws a single completion from the start is not evaluated.
Our evaluation covers mathematical reasoning with six open-weight models from 1.5B to 14B parameters; other task types, such as open-ended generation and multi-turn interaction, are left for future work.
The two-pass policy studied here chooses between two particle budgets; routing across models or reasoning modes with the same signal is a natural extension that we do not evaluate.
\bibliography{custom}

@misc{azizi2026power,
      title={Power-SMC: Low-Latency Sequence-Level Power Sampling for Training-Free LLM Reasoning}, 
      author={Seyedarmin Azizi and Erfan Baghaei Potraghloo and Minoo Ahmadi and Souvik Kundu and Massoud Pedram},
      year={2026},
      eprint={2602.10273},
      archivePrefix={arXiv},
      primaryClass={stat.ML},
      url={https://arxiv.org/abs/2602.10273}, 
}

@book{chopin2020introduction,
  title={An Introduction to Sequential Monte Carlo},
  author={Chopin, N. and Papaspiliopoulos, O.},
  isbn={9783030478445},
  series={Springer Series in Statistics},
  url={https://books.google.com/books?id=dAGAzQEACAAJ},
  year={2020},
  publisher={Springer International Publishing}
}

@book{delmoral2004feynman,
author = {Del Moral, Pierre},
year = {2004},
month = {05},
pages = {},
title = {Feynman-Kac Formulae: Genealogical and Interacting Particle Systems With Applications},
volume = {100},
isbn = {0387202684},
journal = {Journal of The American Statistical Association - J AMER STATIST ASSN},
doi = {10.1007/978-1-4684-9393-1}
}

@article{hendrycks2021measuring,
  title={Measuring Mathematical Problem Solving With the MATH Dataset},
  author={Hendrycks, Dan and Burns, Collin and Kadavath, Saurav and Arora, Akul and Basart, Steven and Tang, Eric and Song, Dawn and Steinhardt, Jacob},
  journal={arXiv preprint arXiv:2103.03874},
  year={2021},
  eprint={2103.03874},
  archivePrefix={arXiv},
  primaryClass={cs.LG},
  url={https://arxiv.org/abs/2103.03874}
}

@article{kadavath2022language,
  title={Language Models (Mostly) Know What They Know},
  author={Saurav Kadavath and Tom Conerly and Amanda Askell and Thomas Henighan and Dawn Drain and Ethan Perez and Nicholas Schiefer and Zachary Dodds and Nova Dassarma and Eli Tran-Johnson and Scott Johnston and Sheer El-Showk and Andy Jones and Nelson Elhage and Tristan Hume and Anna Chen and Yuntao Bai and Sam Bowman and Stanislav Fort and Deep Ganguli and Danny Hernandez and Josh Jacobson and John Kernion and Shauna Kravec and Liane Lovitt and Kamal Ndousse and Catherine Olsson and Sam Ringer and Dario Amodei and Tom B. Brown and Jack Clark and Nicholas Joseph and Benjamin Mann and Sam McCandlish and Chris Olah and Jared Kaplan},
  journal={ArXiv},
  year={2022},
  volume={abs/2207.05221},
  url={https://api.semanticscholar.org/CorpusID:250451161}
}

@inproceedings{
karan2025reasoning,
title={Reasoning with Sampling: Your Base Model is Smarter Than You Think},
author={Aayush Karan and Yilun Du},
booktitle={The Fourteenth International Conference on Learning Representations},
year={2026},
url={https://openreview.net/forum?id=Vsgq2ldr4K}
}

@misc{nguyen2026apps,
      title={The Model Knows, the Decoder Finds: Future Value Guided Particle Power Sampling}, 
      author={Tu Nguyen and Matthieu Zimmer and Rasul Tutunov and Xiaotong Ji and Haitham Bou Ammar},
      year={2026},
      eprint={2605.02427},
      archivePrefix={arXiv},
      primaryClass={cs.AI},
      url={https://arxiv.org/abs/2605.02427}, 
}

@misc{qwen2.5,
      title={Qwen2.5 Technical Report}, 
      author={{Qwen Team} and An Yang and Baosong Yang and Beichen Zhang and Binyuan Hui and Bo Zheng and Bowen Yu and Chengyuan Li and Dayiheng Liu and Fei Huang and Haoran Wei and Huan Lin and Jian Yang and Jianhong Tu and Jianwei Zhang and Jianxin Yang and Jiaxi Yang and Jingren Zhou and Junyang Lin and Kai Dang and Keming Lu and Keqin Bao and Kexin Yang and Le Yu and Mei Li and Mingfeng Xue and Pei Zhang and Qin Zhu and Rui Men and Runji Lin and Tianhao Li and Tianyi Tang and Tingyu Xia and Xingzhang Ren and Xuancheng Ren and Yang Fan and Yang Su and Yichang Zhang and Yu Wan and Yuqiong Liu and Zeyu Cui and Zhenru Zhang and Zihan Qiu},
      year={2025},
      eprint={2412.15115},
      archivePrefix={arXiv},
      primaryClass={cs.CL},
      url={https://arxiv.org/abs/2412.15115}, 
}

@inproceedings{
wang2023selfconsistency,
title={Self-Consistency Improves Chain of Thought Reasoning in Language Models},
author={Xuezhi Wang and Jason Wei and Dale Schuurmans and Quoc V Le and Ed H. Chi and Sharan Narang and Aakanksha Chowdhery and Denny Zhou},
booktitle={The Eleventh International Conference on Learning Representations },
year={2023},
url={https://openreview.net/forum?id=1PL1NIMMrw}
}

@book{cover2006elements,
  title     = {Elements of Information Theory},
  author    = {Cover, Thomas M. and Thomas, Joy A.},
  edition   = {2},
  year      = {2006},
  publisher = {Wiley-Interscience},
  address   = {Hoboken, NJ}
}

@inproceedings{
yue2025does,
title={Does Reinforcement Learning Really Incentivize Reasoning Capacity in {LLM}s Beyond the Base Model?},
author={Yang Yue and Zhiqi Chen and Rui Lu and Andrew Zhao and Zhaokai Wang and Yang Yue and Shiji Song and Gao Huang},
booktitle={The Thirty-ninth Annual Conference on Neural Information Processing Systems},
year={2025},
url={https://openreview.net/forum?id=4OsgYD7em5}
}

@article{Farquhar2024DetectingHI,
  title={Detecting hallucinations in large language models using semantic entropy},
  author={Sebastian Farquhar and Jannik Kossen and Lorenz Kuhn and Yarin Gal},
  journal={Nature},
  year={2024},
  volume={630},
  pages={625 - 630},
  url={https://api.semanticscholar.org/CorpusID:270615909}
}

@misc{cobbe2021training,
      title={Training Verifiers to Solve Math Word Problems}, 
      author={Karl Cobbe and Vineet Kosaraju and Mohammad Bavarian and Mark Chen and Heewoo Jun and Lukasz Kaiser and Matthias Plappert and Jerry Tworek and Jacob Hilton and Reiichiro Nakano and Christopher Hesse and John Schulman},
      year={2021},
      eprint={2110.14168},
      archivePrefix={arXiv},
      primaryClass={cs.LG},
      url={https://arxiv.org/abs/2110.14168}, 
}

@misc{dubey2024llama,
  archiveprefix = {arXiv},
  author = {Dubey, Abhimanyu and Jauhri, Abhinav and Pandey, Abhinav and Kadian, Abhishek and Al-Dahle, Ahmad and Letman, Aiesha and Mathur, Akhil and Schelten, Alan and Yang, Amy and Fan, Angela and Goyal, Anirudh and Hartshorn, Anthony and Yang, Aobo and Mitra, Archi and Sravankumar, Archie and Korenev, Artem and Hinsvark, Arthur and Rao, Arun and Zhang, Aston and Rodriguez, Aurelien and Gregerson, Austen and Spataru, Ava and Roziere, Baptiste and Biron, Bethany and Tang, Binh and Chern, Bobbie and Caucheteux, Charlotte and Nayak, Chaya and Bi, Chloe and Marra, Chris and McConnell, Chris and Keller, Christian and Touret, Christophe and Wu, Chunyang and Wong, Corinne and Ferrer, Cristian Canton and Nikolaidis, Cyrus and Allonsius, Damien and Song, Daniel and Pintz, Danielle and Livshits, Danny and Esiobu, David and Choudhary, Dhruv and Mahajan, Dhruv and Garcia-Olano, Diego and Perino, Diego and Hupkes, Dieuwke and Lakomkin, Egor and AlBadawy, Ehab and Lobanova, Elina and Dinan, Emily and Smith, Eric Michael and Radenovic, Filip and Zhang, Frank and Synnaeve, Gabriel and Lee, Gabrielle and Anderson, Georgia Lewis and Nail, Graeme and Mialon, Gregoire and Pang, Guan and Cucurell, Guillem and Nguyen, Hailey and Korevaar, Hannah and Xu, Hu and Touvron, Hugo and Zarov, Iliyan and Ibarra, Imanol Arrieta and Kloumann, Isabel and Misra, Ishan and Evtimov, Ivan and Copet, Jade and Lee, Jaewon and Geffert, Jan and Vranes, Jana and Park, Jason and Mahadeokar, Jay and Shah, Jeet and van der Linde, Jelmer and Billock, Jennifer and Hong, Jenny and Lee, Jenya and Fu, Jeremy and Chi, Jianfeng and Huang, Jianyu and Liu, Jiawen and Wang, Jie and Yu, Jiecao and Bitton, Joanna and Spisak, Joe and Park, Jongsoo and Rocca, Joseph and Johnstun, Joshua and Saxe, Joshua and Jia, Junteng and Alwala, Kalyan Vasuden and Upasani, Kartikeya and Plawiak, Kate and Li, Ke and Heafield, Kenneth and Stone, Kevin and El-Arini, Khalid and Iyer, Krithika and Malik, Kshitiz and Chiu, Kuenley and Bhalla, Kunal and Rantala-Yeary, Lauren and van der Maaten, Laurens and Chen, Lawrence and Tan, Liang and Jenkins, Liz and Martin, Louis and Madaan, Lovish and Malo, Lubo and Blecher, Lukas and Landzaat, Lukas and de Oliveira, Luke and Muzzi, Madeline and Pasupuleti, Mahesh and Singh, Mannat and Paluri, Manohar and Kardas, Marcin and Oldham, Mathew and Rita, Mathieu and Pavlova, Maya and Kambadur, Melanie and Lewis, Mike and Si, Min and Singh, Mitesh Kumar and Hassan, Mona and Goyal, Naman and Torabi, Narjes and Bashlykov, Nikolay and Bogoychev, Nikolay and Chatterji, Niladri and Duchenne, Olivier and Çelebi, Onur and Alrassy, Patrick and Zhang, Pengchuan and Li, Pengwei and Vasic, Petar and Weng, Peter and Bhargava, Prajjwal and Dubal, Pratik and Krishnan, Praveen and Koura, Punit Singh and Xu, Puxin and He, Qing and Dong, Qingxiao and Srinivasan, Ragavan and Ganapathy, Raj and Calderer, Ramon and Cabral, Ricardo Silveira and Stojnic, Robert and Raileanu, Roberta and Girdhar, Rohit and Patel, Rohit and Sauvestre, Romain and Polidoro, Ronnie and Sumbaly, Roshan and Taylor, Ross and Silva, Ruan and Hou, Rui and Wang, Rui and Hosseini, Saghar and Chennabasappa, Sahana and Singh, Sanjay and Bell, Sean and Kim, Seohyun Sonia and Edunov, Sergey and Nie, Shaoliang and Narang, Sharan and Raparthy, Sharath and Shen, Sheng and Wan, Shengye and Bhosale, Shruti and Zhang, Shun and Vandenhende, Simon and Batra, Soumya and Whitman, Spencer and Sootla, Sten and Collot, Stephane and Gururangan, Suchin and Borodinsky, Sydney and Herman, Tamar and Fowler, Tara and Sheasha, Tarek and Georgiou, Thomas and Scialom, Thomas and Speckbacher, Tobias and Mihaylov, Todor and Xiao, Tong and Karn, Ujjwal and Goswami, Vedanuj and Gupta, Vibhor and Ramanathan, Vignesh and Kerkez, Viktor and Gonguet, Vincent and Do, Virginie and Vogeti, Vish and Petrovic, Vladan and Chu, Weiwei and Xiong, Wenhan and Fu, Wenyin and Meers, Whitney and Martinet, Xavier and Wang, Xiaodong and Tan, Xiaoqing Ellen and Xie, Xinfeng and Jia, Xuchao and Wang, Xuewei and Goldschlag, Yaelle and Gaur, Yashesh and Babaei, Yasmine and Wen, Yi and Song, Yiwen and Zhang, Yuchen and Li, Yue and Mao, Yuning and Coudert, Zacharie Delpierre and Yan, Zheng and Chen, Zhengxing and Papakipos, Zoe and Singh, Aaditya and Grattafiori, Aaron and Jain, Abha and Kelsey, Adam and Shajnfeld, Adam and Gangidi, Adithya and Victoria, Adolfo and Goldstand, Ahuva and Menon, Ajay and Sharma, Ajay and Boesenberg, Alex and Vaughan, Alex and Baevski, Alexei and Feinstein, Allie and Kallet, Amanda and Sangani, Amit and Yunus, Anam and Lupu, Andrei and Alvarado, Andres and Caples, Andrew and Gu, Andrew and Ho, Andrew and Poulton, Andrew and Ryan, Andrew and Ramchandani, Ankit and Franco, Annie and Saraf, Aparajita and Chowdhury, Arkabandhu and Gabriel, Ashley and Bharambe, Ashwin and Eisenman, Assaf and Yazdan, Azadeh and James, Beau and Maurer, Ben and Leonhardi, Benjamin and Huang, Bernie and Loyd, Beth and Paola, Beto De and Paranjape, Bhargavi and Liu, Bing and Wu, Bo and Ni, Boyu and Hancock, Braden and Wasti, Bram and Spence, Brandon and Stojkovic, Brani and Gamido, Brian and Montalvo, Britt and Parker, Carl and Burton, Carly and Mejia, Catalina and Wang, Changhan and Kim, Changkyu and Zhou, Chao and Hu, Chester and Chu, Ching-Hsiang and Cai, Chris and Tindal, Chris and Feichtenhofer, Christoph and Civin, Damon and Beaty, Dana and Kreymer, Daniel and Li, Daniel and Wyatt, Danny and Adkins, David and Xu, David and Testuggine, Davide and David, Delia and Parikh, Devi and Liskovich, Diana and Foss, Didem and Wang, Dingkang and Le, Duc and Holland, Dustin and Dowling, Edward and Jamil, Eissa and Montgomery, Elaine and Presani, Eleonora and Hahn, Emily and Wood, Emily and Brinkman, Erik and Arcaute, Esteban and Dunbar, Evan and Smothers, Evan and Sun, Fei and Kreuk, Felix and Tian, Feng and Ozgenel, Firat and Caggioni, Francesco and Guzmán, Francisco and Kanayet, Frank and Seide, Frank and Florez, Gabriela Medina and Schwarz, Gabriella and Badeer, Gada and Swee, Georgia and Halpern, Gil and Thattai, Govind and Herman, Grant and Sizov, Grigory and Guangyi and Zhang and Lakshminarayanan, Guna and Shojanazeri, Hamid and Zou, Han and Wang, Hannah and Zha, Hanwen and Habeeb, Haroun and Rudolph, Harrison and Suk, Helen and Aspegren, Henry and Goldman, Hunter and Damlaj, Ibrahim and Molybog, Igor and Tufanov, Igor and Veliche, Irina-Elena and Gat, Itai and Weissman, Jake and Geboski, James and Kohli, James and Asher, Japhet and Gaya, Jean-Baptiste and Marcus, Jeff and Tang, Jeff and Chan, Jennifer and Zhen, Jenny and Reizenstein, Jeremy and Teboul, Jeremy and Zhong, Jessica and Jin, Jian and Yang, Jingyi and Cummings, Joe and Carvill, Jon and Shepard, Jon and McPhie, Jonathan and Torres, Jonathan and Ginsburg, Josh and Wang, Junjie and Wu, Kai and U, Kam Hou and Saxena, Karan and Prasad, Karthik and Khandelwal, Kartikay and Zand, Katayoun and Matosich, Kathy and Veeraraghavan, Kaushik and Michelena, Kelly and Li, Keqian and Huang, Kun and Chawla, Kunal and Lakhotia, Kushal and Huang, Kyle and Chen, Lailin and Garg, Lakshya and A, Lavender and Silva, Leandro and Bell, Lee and Zhang, Lei and Guo, Liangpeng and Yu, Licheng and Moshkovich, Liron and Wehrstedt, Luca and Khabsa, Madian and Avalani, Manav and Bhatt, Manish and Tsimpoukelli, Maria and Mankus, Martynas and Hasson, Matan and Lennie, Matthew and Reso, Matthias and Groshev, Maxim and Naumov, Maxim and Lathi, Maya and Keneally, Meghan and Seltzer, Michael L. and Valko, Michal and Restrepo, Michelle and Patel, Mihir and Vyatskov, Mik and Samvelyan, Mikayel and Clark, Mike and Macey, Mike and Wang, Mike and Hermoso, Miquel Jubert and Metanat, Mo and Rastegari, Mohammad and Bansal, Munish and Santhanam, Nandhini and Parks, Natascha and White, Natasha and Bawa, Navyata and Singhal, Nayan and Egebo, Nick and Usunier, Nicolas and Laptev, Nikolay Pavlovich and Dong, Ning and Zhang, Ning and Cheng, Norman and Chernoguz, Oleg and Hart, Olivia and Salpekar, Omkar and Kalinli, Ozlem and Kent, Parkin and Parekh, Parth and Saab, Paul and Balaji, Pavan and Rittner, Pedro and Bontrager, Philip and Roux, Pierre and Dollar, Piotr and Zvyagina, Polina and Ratanchandani, Prashant and Yuvraj, Pritish and Liang, Qian and Alao, Rachad and Rodriguez, Rachel and Ayub, Rafi and Murthy, Raghotham and Nayani, Raghu and Mitra, Rahul and Li, Raymond and Hogan, Rebekkah and Battey, Robin and Wang, Rocky and Maheswari, Rohan and Howes, Russ and Rinott, Ruty and Bondu, Sai Jayesh and Datta, Samyak and Chugh, Sara and Hunt, Sara and Dhillon, Sargun and Sidorov, Sasha and Pan, Satadru and Verma, Saurabh and Yamamoto, Seiji and Ramaswamy, Sharadh and Lindsay, Shaun and Lindsay, Shaun and Feng, Sheng and Lin, Shenghao and Zha, Shengxin Cindy and Shankar, Shiva and Zhang, Shuqiang and Zhang, Shuqiang and Wang, Sinong and Agarwal, Sneha and Sajuyigbe, Soji and Chintala, Soumith and Max, Stephanie and Chen, Stephen and Kehoe, Steve and Satterfield, Steve and Govindaprasad, Sudarshan and Gupta, Sumit and Cho, Sungmin and Virk, Sunny and Subramanian, Suraj and Choudhury, Sy and Goldman, Sydney and Remez, Tal and Glaser, Tamar and Best, Tamara and Kohler, Thilo and Robinson, Thomas and Li, Tianhe and Zhang, Tianjun and Matthews, Tim and Chou, Timothy and Shaked, Tzook and Vontimitta, Varun and Ajayi, Victoria and Montanez, Victoria and Mohan, Vijai and Kumar, Vinay Satish and Mangla, Vishal and Albiero, Vítor and Ionescu, Vlad and Poenaru, Vlad and Mihailescu, Vlad Tiberiu and Ivanov, Vladimir and Li, Wei and Wang, Wenchen and Jiang, Wenwen and Bouaziz, Wes and Constable, Will and Tang, Xiaocheng and Wang, Xiaofang and Wu, Xiaojian and Wang, Xiaolan and Xia, Xide and Wu, Xilun and Gao, Xinbo and Chen, Yanjun and Hu, Ye and Jia, Ye and Qi, Ye and Li, Yenda and Zhang, Yilin and Zhang, Ying and Adi, Yossi and Nam, Youngjin and Yu and Wang and Hao, Yuchen and Qian, Yundi and He, Yuzi and Rait, Zach and DeVito, Zachary and Rosnbrick, Zef and Wen, Zhaoduo and Yang, Zhenyu and Zhao, Zhiwei},
  description = {The Llama 3 Herd of Models},
  eprint = {2407.21783},
  primaryclass = {cs.AI},
  title = {The Llama 3 Herd of Models},
  url = {https://arxiv.org/abs/2407.21783},
  year = 2024
}

@misc{kuhn2023semantic,
      title={Semantic Uncertainty: Linguistic Invariances for Uncertainty Estimation in Natural Language Generation}, 
      author={Lorenz Kuhn and Yarin Gal and Sebastian Farquhar},
      year={2023},
      eprint={2302.09664},
      archivePrefix={arXiv},
      primaryClass={cs.CL},
      url={https://arxiv.org/abs/2302.09664}, 
}

@misc{ji2026scalable,
      title={Scalable Power Sampling: Unlocking Efficient, Training-Free Reasoning for LLMs via Distribution Sharpening}, 
      author={Xiaotong Ji and Rasul Tutunov and Matthieu Zimmer and Haitham Bou Ammar},
      year={2026},
      eprint={2601.21590},
      archivePrefix={arXiv},
      primaryClass={cs.LG},
      url={https://arxiv.org/abs/2601.21590}, 
}

@misc{snell2024scaling,
      title={Scaling LLM Test-Time Compute Optimally can be More Effective than Scaling Model Parameters}, 
      author={Charlie Snell and Jaehoon Lee and Kelvin Xu and Aviral Kumar},
      year={2024},
      eprint={2408.03314},
      archivePrefix={arXiv},
      primaryClass={cs.LG},
      url={https://arxiv.org/abs/2408.03314}, 
}

@inproceedings{fu2025certaindex,
  title={Efficiently Scaling {LLM} Reasoning Programs with Certaindex},
  author={Fu, Yichao and Chen, Junda and Zhu, Siqi and Fu, Zheyu and Dai, Zhongdongming and Zhuang, Yonghao and Ma, Yian and Qiao, Aurick and Rosing, Tajana and Stoica, Ion and Zhang, Hao},
  booktitle={Advances in Neural Information Processing Systems},
  year={2025}
}

@inproceedings{kang2025selfcertainty,
  title={Scalable Best-of-N Selection for Large Language Models via Self-Certainty},
  author={Kang, Zhewei and Zhao, Xuandong and Song, Dawn},
  booktitle={Advances in Neural Information Processing Systems},
  year={2025}
}

@inproceedings{huang2026cats,
  title={{CaTS}: Calibrated Test-Time Scaling for Efficient {LLM} Reasoning},
  author={Huang, Chengsong and Huang, Langlin and Leng, Jixuan and Liu, Jiacheng and Huang, Jiaxin},
  booktitle={International Conference on Learning Representations},
  year={2026}
}

@misc{yang2025qwen3,
  title={Qwen3 Technical Report},
  author={Yang, An and others},
  year={2025},
  eprint={2505.09388},
  archivePrefix={arXiv},
  primaryClass={cs.CL}
}

@inproceedings{malinin2021uncertainty,
  title={Uncertainty Estimation in Autoregressive Structured Prediction},
  author={Malinin, Andrey and Gales, Mark},
  booktitle={International Conference on Learning Representations},
  year={2021}
}

@article{arikan1996inequality,
  title={An inequality on guessing and its application to sequential decoding},
  author={Arikan, Erdal},
  journal={IEEE Transactions on Information Theory},
  volume={42},
  number={1},
  pages={99--105},
  year={1996}
}

@inproceedings{darrin2023rainproof,
  title={{RainProof}: An Umbrella to Shield Text Generator from Out-Of-Distribution Data},
  author={Darrin, Maxime and Piantanida, Pablo and Colombo, Pierre},
  booktitle={Proceedings of the 2023 Conference on Empirical Methods in Natural Language Processing},
  pages={5831--5857},
  year={2023}
}

@article{lightman2023verify,
  title={Let's Verify Step by Step},
  author={Lightman, Hunter and Kosaraju, Vineet and Burda, Yura and Edwards, Harri and Baker, Bowen and Lee, Teddy and Leike, Jan and Schulman, John and Sutskever, Ilya and Cobbe, Karl},
  journal={arXiv preprint arXiv:2305.20050},
  year={2023}
}

@article{jizba2004world,
  title={The world according to {R}{\'e}nyi: thermodynamics of multifractal systems},
  author={Jizba, Petr and Arimitsu, Toshihico},
  journal={Annals of Physics},
  volume={312},
  number={1},
  pages={17--59},
  year={2004}
}

\clearpage
\appendix

\section{Power-SMC Accuracy}
\label{app:smcbaseline}

Table~\ref{tab:model_acc} reports the accuracy of Power-SMC for the four models on \textbf{MATH500} \citep{hendrycks2021measuring} and \textbf{GSM8K} \citep{cobbe2021training}.
All models are decoded with Power-SMC ($\alpha{=}4$, block size 64, ESS threshold 0.5, $\alpha$-ramp over 400 tokens, top-$p$ 0.9 on the proposal, at most 2{,}048 new tokens) unless noted otherwise.
The default particle budget is $N{=}32$.
Answers are graded with the Power-SMC repository's exact-match grader.

\begin{table}[h]
\centering
\small
\begin{tabular}{@{}lcc@{}}
\toprule
\textbf{Model} & \textbf{MATH500} & \textbf{GSM8K} \\
\midrule
Qwen2.5-Math-1.5B & 69.0 & 80.8 \\
Qwen2.5-Math-7B & 75.8 & 86.2 \\
Qwen2.5-7B & 71.8 & 85.0 \\
Llama-3.1-8B-Instruct & 50.2 & 85.2 \\
\bottomrule
\end{tabular}
\caption{Model accuracy (\%) with Power-SMC ($N{=}32$, $\alpha{=}4$, 500 problems).}
\label{tab:model_acc}
\end{table}
\begin{table}[h]
\centering
\small
\begin{tabular}{@{}crccc@{}}
\toprule
\textbf{Lvl} & $n$ & $\Hord$ & \textbf{TokEnt} & \textbf{Acc} \\
\midrule
1 & 43 & 12.8 & 0.088 & 0.91 \\
2 & 90 & 16.4 & 0.077 & 0.87 \\
3 & 105 & 20.8 & 0.089 & 0.87 \\
4 & 128 & 24.2 & 0.085 & 0.73 \\
5 & 134 & 32.6 & 0.090 & 0.57 \\
\bottomrule
\end{tabular}
\caption{Per-level mean uncertainty and accuracy (500 problems, Qwen2.5-Math-7B).}
\label{tab:perlevel}
\end{table}
\begin{table}[h]
\centering
\small
\setlength{\tabcolsep}{4pt}
\begin{tabular}{@{}lcccc@{}}
\toprule
\textbf{Signal} & \textbf{M-1.5B} & \textbf{M-7B} & \textbf{Q-7B} & \textbf{Llama} \\
\midrule
$\Hone$ (ours) & +0.461 & \textbf{+0.447} & +0.480 & +0.499 \\
$\Hord$ (ours) & +0.467 & +0.441 & +0.494 & +0.500 \\
Neg. seq log-prob & \textbf{+0.473} & +0.436 & \textbf{+0.502} & \textbf{+0.505} \\
Mean token entropy & +0.164 & +0.094 & $-$0.120 & $-$0.120 \\
\bottomrule
\end{tabular}
\caption{Spearman $\rho$ with MATH difficulty levels (all 500 problems, $N{=}32$, $\alpha{=}4$).}
\label{tab:difficulty}
\end{table}

\paragraph{Difficulty estimation.}
We ask whether uncertainty signals track externally annotated problem difficulty.
MATH500 problems have ground-truth difficulty levels from 1 (easiest) to 5 (hardest) \citep{hendrycks2021measuring}.
We compute the Spearman rank correlation between each signal and the level over all 500 problems (Table~\ref{tab:difficulty}).

$\Hone$ and $\Hord$ correlate with difficulty at $\rho{=}0.44$ to $0.50$ across the four models, on par with sequence log-probability, while mean token entropy reaches at most 0.16 and is negatively correlated for two models.
Per-level means for Qwen2.5-Math-7B (Table~\ref{tab:perlevel}) confirm the trend: $\Hord$ increases monotonically from 12.8 (Level~1) to 32.6 (Level~5), while accuracy decreases from 0.91 to 0.57.
The same monotone increase holds for the other three models.

\section{Correct vs.\ Incorrect Signal Distributions}
\label{app:density}

\begin{figure}[h]
  \centering
  \includegraphics[width=\columnwidth]{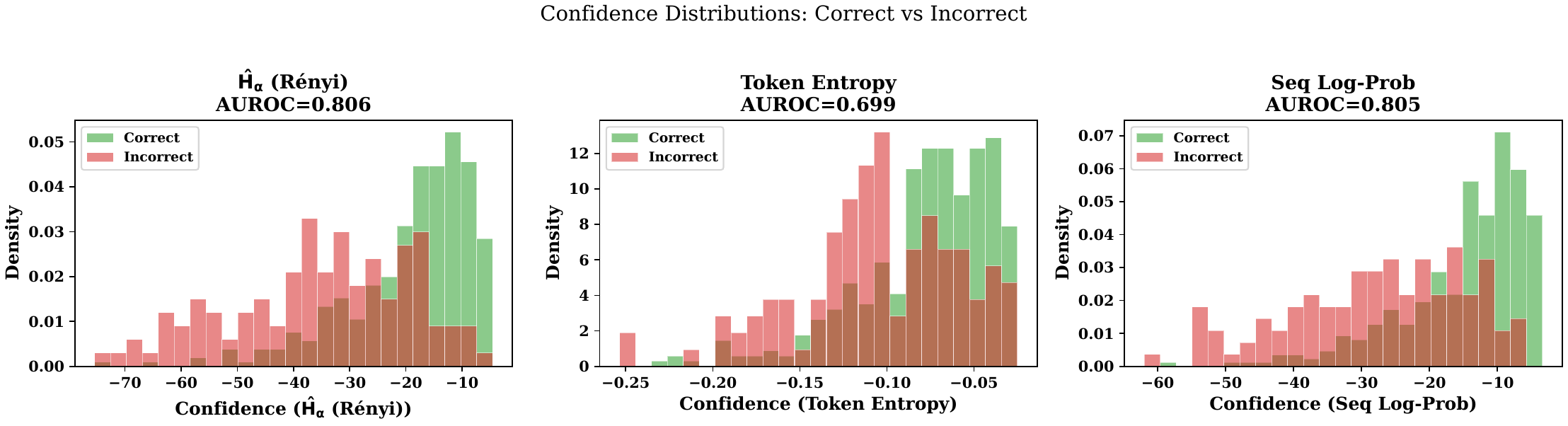}
  \caption{Confidence distributions for correct (green) and incorrect (red) answers under each signal.}
  \label{fig:density}
\end{figure}

Figure~\ref{fig:density} shows the distributions of each uncertainty signal for correct and incorrect answers (MATH500, Qwen2.5-Math-7B).
$\Hord$ separates the two groups (AUROC 0.806), with incorrect answers concentrating at high-uncertainty values; $\Hhalf$ reaches 0.850.
Mean token entropy (AUROC 0.699) shows more overlap.

\section{Particle Agreement}
\label{app:sc}

Particle agreement (Self-cons.\ in Table~\ref{tab:main}) measures answer agreement across the $N{=}32$ SMC particles.
After resampling, particles share ancestors and are no longer independent, which suppresses the disagreements that signal incorrect answers.
Weighting the particles' answers by the normalized weights does not fix particle agreement: the weighted answer entropy reaches 0.60 to 0.72 AUROC in the eight $N{=}8$ and $N{=}64$ runs of \S\ref{sec:pareto}, below sequence log-probability (0.78 to 0.81) in each.
Computed along the returned completion alone, the order-1/2 sum is above sequence log-probability in all seven settings where we log it, by 0.013 to 0.033 AUROC.
\section{Signal Combination}
\label{app:combination}
A natural question is whether the four zero-cost signals capture complementary information.
We train a logistic regression (5-fold cross-validated) on MATH500 with Qwen2.5-Math-7B to predict correctness from signal combinations.
$\Hhalf$ alone reaches 0.844 cross-validated AUROC; adding all three other signals raises this to 0.878.
Adding $\Hhalf$ to the three baseline signals raises the cross-validated AUROC in all eight settings of Table~\ref{tab:main}, by 0.001 to 0.023.
The improvement indicates that the signals capture partially orthogonal aspects of uncertainty, and a simple learned combination outperforms any single signal.
In practice, this combination remains zero-cost at inference: all four signals are already computed by the SMC run.

\section{Positioning Against Prior Work}
\label{app:positioning}

\begin{table}[h]
\centering
\small
\begin{tabular}{@{}p{2.2cm}p{1.8cm}p{1.9cm}@{}}
\toprule
\textbf{Method} & \textbf{Level} & \textbf{Extra cost} \\
\midrule
Token ent. & Token & Free \\
Seq log-prob & Single path & Free \\
Self-Certainty & Token dist. & Free \\
Self-cons. & Answer & $K$ gens \\
Certaindex & Answer & $K$ gens \\
Sem.\ ent. & Ans.\ clust. & $K$ + clust. \\
P(True) & Surface & 1 query \\
\midrule
SRE (ours) & \textbf{Seq.\ dist.} & \textbf{0 marg.} \\
\bottomrule
\end{tabular}
\caption{Comparison with LLM uncertainty methods. Level: the object the signal is computed on. Extra cost: computation beyond the decoding run that produces the answer.}
\label{tab:positioning}
\end{table}
\section{Handling Resampling in the Normalizing-Constant Estimate}
\label{app:resampling}

Power-SMC resamples at block boundaries when the effective sample size falls below a threshold, so between resampling events the particles carry unequal normalized weights $W_{t-1}^{(i)}$.
The standard estimator of Eq.~\ref{eq:zhat} accumulates the log of the $W$-weighted mean of the incremental weights at every step.
Over a segment between two resampling events these terms telescope to $\log \frac{1}{N}\sum_i \widetilde{w}^{(i)}$, where $\widetilde{w}^{(i)}$ is the unnormalized weight that particle $i$ has accumulated since the last resampling event.
When systematic resampling occurs, the weights are reset to $1/N$ and the unbiasedness of $\Zalpha$ is preserved because systematic resampling is itself unbiased \citep{chopin2020introduction}: $\mathbb{E}[\frac{1}{N}\sum_i f(y^{(A_i)})] = \sum_i w_i f(y^{(i)})$.
The stepwise accumulation of Eq.~\ref{eq:zstep} uses the uniform mean at every step, so it needs no bookkeeping at resampling boundaries: we continue accumulating log-mean incremental weights across them.
Power-SMC raises the exponent linearly from 1 to $\alpha$ over the first 400 tokens (the exponent ramp), $\alpha_t = 1 + (\alpha-1)\min(t,400)/400$ with $\alpha_0 = 1$.
The intermediate target after $t$ tokens is $\ptheta(y_{1:t}|x)^{\alpha_t}$ and the incremental weight is
\begin{align*}
\omega_t = {}& \ptheta(y_{<t}|x)^{\alpha_t-\alpha_{t-1}}\,\ptheta(y_t|x,y_{<t})^{\alpha_t-\alpha} \\
& \times \textstyle\sum_{v\in V_t}\ptheta(v|x,y_{<t})^{\alpha},
\end{align*}
where $V_t$ is the nucleus of the proposal (the tokens that nucleus truncation keeps).
The product of the incremental weights along a completion equals $\ptheta(y|x)^\alpha/q(y)$ for every completion length, because a final correction raises the exponent to $\alpha$ when decoding ends before the ramp does.
The final target is therefore $\pi_\alpha$ and Eq.~\ref{eq:zhat} estimates $Z_\alpha$.
Both accumulations use this incremental weight.

\begin{algorithm}[h]
\small
\caption{SRE extraction from Power-SMC}
\label{alg:extract}
\begin{algorithmic}[1]
\State $\texttt{log\_Z} \gets 0$ \hfill \textit{// before decoding}
\For{$t = 1,\ldots,T$}
  \State \textit{// existing Power-SMC proposal, sampling, weighting}
  \State \textit{// $\omega_t^{(1)},\ldots,\omega_t^{(N)}$ already computed}
  \State $\texttt{log\_Z}
      \mathrel{+}{=}
      \log\!\left(\frac{1}{N}\sum_i \omega_t^{(i)}\right)$
\EndFor
\State SRE $ \gets \texttt{log\_Z}/(1-\alpha)$
\end{algorithmic}
\end{algorithm}

\section{Estimator Variants}
\label{app:variants}

Table~\ref{tab:variants} compares the stepwise accumulation of Eq.~\ref{eq:zstep}, reported as $\Hord$ throughout the paper, with the standard estimator of Eq.~\ref{eq:zhat} and with $\Hone$, all computed in the same runs.
The stepwise and standard estimators rank problems almost identically (Spearman $\rho \geq 0.98$ in every row).
The standard estimator stays within 0.004 AUROC of sequence log-probability at $\alpha{=}4$, while the stepwise accumulation is above log-probability in all 13 rows.
The order-$\alpha$ member of Eq.~\ref{eq:family} (Fam.), which uses no ramp term and averages over the particles still generating, agrees with the stepwise accumulation at Spearman $\rho \geq 0.94$.
As a gating signal for adaptive particle allocation (\S\ref{sec:pareto}), the stepwise accumulation reaches uniform $N{=}64$ accuracy with 47.5\% fewer particles on the first seed and the standard estimator with 42.5\%.

\begin{table}[h]
\centering
\small
\setlength{\tabcolsep}{2pt}
\begin{tabular}{@{}llcccccc@{}}
\toprule
 & \textbf{Model, $\alpha$} & \textbf{LogP} & \textbf{Std.} & \textbf{Step.} & \textbf{Fam.} & $\Hone$ & $\Hhalf$ \\
\midrule
MATH & M-1.5B, 4 & .810 & .810 & .816 & .818 & .825 & .836 \\
MATH & M-7B, 4 & .805 & .805 & .806 & .816 & .821 & .850 \\
MATH & Q-7B, 4 & .801 & .804 & .809 & .816 & .824 & .834 \\
MATH & Llama, 4 & .862 & .862 & .866 & .858 & .866 & .884 \\
GSM & M-1.5B, 4 & .744 & .745 & .748 & .749 & .760 & .776 \\
GSM & M-7B, 4 & .741 & .745 & .753 & .771 & .779 & .782 \\
GSM & Q-7B, 4 & .635 & .636 & .642 & .685 & .686 & .658 \\
GSM & Llama, 4 & .719 & .723 & .724 & .735 & .733 & .743 \\
\midrule
MATH & M-7B, 1.5 & .730 & .740 & .772 & .800 & .802 & .808 \\
MATH & M-7B, 2 & .772 & .772 & .790 & .810 & .814 & .841 \\
MATH & M-7B, 3 & .782 & .782 & .796 & .810 & .813 & .828 \\
MATH & M-7B, 6 & .786 & .787 & .798 & .801 & .807 & .841 \\
MATH & M-7B, 8 & .803 & .803 & .810 & .807 & .822 & .849 \\
\bottomrule
\end{tabular}
\caption{Selective prediction AUROC ($N{=}32$) of sequence log-probability (LogP), the standard estimator (Std., Eq.~\ref{eq:zhat}), the stepwise accumulation of order $\alpha$ (Step., Eq.~\ref{eq:zstep}), and the members of Eq.~\ref{eq:family} of order $\alpha$ (Fam.), 1, and 1/2.}
\label{tab:variants}
\end{table}

\section{Problem Transitions Under Particle Scaling}
\label{app:transitions}

Table~\ref{tab:transitions} characterizes how individual problems change when scaling from $N{=}8$ to $N{=}64$ (500 problems).
The stepwise accumulation from the probe separates the problems that are answered correctly at both particle budgets (mean $\Hord$ 19.3) from the 44 whose correctness changes (33.4) and the 98 that stay wrong (41.1).
Probe $\Hhalf$ identifies the problems that stay correct with AUROC 0.841, 0.845, and 0.845 on the three seeds, against 0.809, 0.818, and 0.808 for log-probability.

\begin{table}[h]
\centering
\small
\begin{tabular}{@{}lrc@{}}
\toprule
\textbf{Transition} & $n$ & \textbf{Mean probe $\Hord$} \\
\midrule
Stayed correct & 358 & 19.3 \\
Flipped to correct & 28 & 34.2 \\
Flipped to wrong & 16 & 31.9 \\
Stayed wrong & 98 & 41.1 \\
\bottomrule
\end{tabular}
\caption{Problem transitions $N{=}8 \to N{=}64$ (500 problems, Qwen2.5-Math-7B).}
\label{tab:transitions}
\end{table}

\section{Bootstrap Confidence Intervals}
\label{app:bootstrap}

Tables~\ref{tab:bootstrap} and~\ref{tab:bootstrap_gsm} report 95\% bootstrap confidence intervals (10{,}000 replicates) and paired permutation test $p$-values against sequence log-probability (10{,}000 permutations; 200{,}000 for the one value within 0.005 of 0.05) for the eight settings of Table~\ref{tab:main}.

\begin{table}[h]
\centering
\small
\setlength{\tabcolsep}{3pt}
\begin{tabular}{@{}llcc@{}}
\toprule
\textbf{Model} & \textbf{Signal} & \textbf{AUROC [95\% CI]} & $p$ \textbf{vs.}\ \textbf{LogP} \\
\midrule
\multirow{6}{*}{M-1.5B}
  & $\Hhalf$ & .836 [.797, .871] & .002$^{**}$ \\
  & $\Hone$ & .825 [.784, .862] & .007$^{**}$ \\
  & $\Hord$ & .816 [.774, .854] & .035$^{*}$ \\
  & Seq log-prob & .810 [.767, .848] & --- \\
  & Self-cons.$^\dag$ & .656 [.613, .699] & ${<}.001^{***}$ \\
  & Token entropy & .714 [.663, .764] & ${<}.001^{***}$ \\
\midrule
\multirow{6}{*}{M-7B}
  & $\Hhalf$ & .850 [.808, .887] & ${<}.001^{***}$ \\
  & $\Hone$ & .821 [.776, .862] & .013$^{*}$ \\
  & $\Hord$ & .806 [.759, .848] & .892 \\
  & Seq log-prob & .805 [.760, .848] & --- \\
  & Self-cons.$^\dag$ & .696 [.647, .744] & .003$^{**}$ \\
  & Token entropy & .699 [.643, .752] & ${<}.001^{***}$ \\
\midrule
\multirow{6}{*}{Q-7B}
  & $\Hhalf$ & .834 [.795, .872] & .003$^{**}$ \\
  & $\Hone$ & .824 [.783, .863] & .005$^{**}$ \\
  & $\Hord$ & .809 [.766, .851] & .121 \\
  & Seq log-prob & .801 [.756, .845] & --- \\
  & Self-cons.$^\dag$ & .650 [.604, .695] & ${<}.001^{***}$ \\
  & Token entropy & .531 [.470, .590] & ${<}.001^{***}$ \\
\midrule
\multirow{6}{*}{Llama}
  & $\Hhalf$ & .884 [.852, .913] & .006$^{**}$ \\
  & $\Hone$ & .866 [.832, .897] & .461 \\
  & $\Hord$ & .866 [.832, .897] & .090 \\
  & Seq log-prob & .862 [.827, .893] & --- \\
  & Self-cons.$^\dag$ & .547 [.511, .583] & ${<}.001^{***}$ \\
  & Token entropy & .390 [.341, .440] & ${<}.001^{***}$ \\
\bottomrule
\multicolumn{4}{@{}l}{\scriptsize $^\dag$Correlated SMC-particle agreement.}
\end{tabular}
\caption{Bootstrap 95\% confidence intervals and paired permutation tests on MATH500 ($\alpha{=}4$, $N{=}32$).}
\label{tab:bootstrap}
\end{table}

\begin{table}[h]
\centering
\small
\setlength{\tabcolsep}{3pt}
\begin{tabular}{@{}llcc@{}}
\toprule
\textbf{Model} & \textbf{Signal} & \textbf{AUROC [95\% CI]} & $p$ \textbf{vs.}\ \textbf{LogP} \\
\midrule
\multirow{6}{*}{M-1.5B}
  & $\Hhalf$ & .776 [.725, .823] & .020$^{*}$ \\
  & $\Hone$ & .760 [.706, .811] & .049$^{*}$ \\
  & $\Hord$ & .748 [.693, .800] & .251 \\
  & Seq log-prob & .744 [.689, .795] & --- \\
  & Self-cons.$^\dag$ & .718 [.664, .770] & .575 \\
  & Token entropy & .728 [.666, .788] & .519 \\
\midrule
\multirow{6}{*}{M-7B}
  & $\Hhalf$ & .782 [.723, .837] & .063 \\
  & $\Hone$ & .779 [.725, .830] & .002$^{**}$ \\
  & $\Hord$ & .753 [.697, .806] & .014$^{*}$ \\
  & Seq log-prob & .741 [.683, .795] & --- \\
  & Self-cons.$^\dag$ & .713 [.648, .776] & .500 \\
  & Token entropy & .820 [.766, .869] & .005$^{**}$ \\
\midrule
\multirow{6}{*}{Q-7B}
  & $\Hhalf$ & .658 [.586, .726] & .213 \\
  & $\Hone$ & .686 [.619, .752] & ${<}.001^{***}$ \\
  & $\Hord$ & .642 [.566, .717] & .210 \\
  & Seq log-prob & .635 [.559, .710] & --- \\
  & Self-cons.$^\dag$ & .688 [.628, .746] & .325 \\
  & Token entropy & .697 [.631, .763] & .123 \\
\midrule
\multirow{6}{*}{Llama}
  & $\Hhalf$ & .743 [.674, .806] & .129 \\
  & $\Hone$ & .733 [.662, .797] & .255 \\
  & $\Hord$ & .724 [.651, .791] & .419 \\
  & Seq log-prob & .719 [.646, .787] & --- \\
  & Self-cons.$^\dag$ & .626 [.558, .691] & .094 \\
  & Token entropy & .691 [.624, .756] & .346 \\
\bottomrule
\multicolumn{4}{@{}l}{\scriptsize $^\dag$Correlated SMC-particle agreement.}
\end{tabular}
\caption{Bootstrap 95\% confidence intervals and paired permutation tests on GSM8K ($\alpha{=}4$, $N{=}32$).}
\label{tab:bootstrap_gsm}
\end{table}

$\Hhalf$ is above sequence log-probability in all eight settings, significantly for all four models on MATH500 ($p{=}0.002$, ${<}0.001$, $0.003$, and $0.006$) and for Qwen2.5-Math-1.5B on GSM8K ($p{=}0.020$).
With Holm correction over the eight settings it remains significant for all four MATH500 models (adjusted $p \leq 0.029$).
It is significantly above $\Hone$ for Qwen2.5-Math-7B and Llama on MATH500 ($p < 0.001$), and significantly above generation length in all eight settings ($p \leq 0.001$ after Holm correction).
$\Hone$ is above sequence log-probability in all eight settings, significantly for the three Qwen models on both benchmarks, and $\Hord$ is above it in all eight, significantly in two.
Mean token entropy and particle agreement are significantly below log-probability for all four models on MATH500 ($p \leq 0.003$).

\section{Larger and Newer Models}
\label{app:larger}

Table~\ref{tab:larger} repeats the selective-prediction experiment on MATH500 with Qwen2.5-14B-Instruct and with Qwen3-4B \citep{yang2025qwen3} in non-thinking mode, using the same Power-SMC hyperparameters as in Table~\ref{tab:main}.
Power-SMC reaches 80.8\% and 79.8\% accuracy.
All three SRE signals are above sequence log-probability with both models.
With Qwen3-4B, $\Hhalf$ is the best signal, 0.026 above log-probability ($p < 0.001$) and 0.014 above the same sum along the returned completion ($p{=}0.005$).

\begin{table}[h]
\centering
\small
\begin{tabular}{@{}lcc@{}}
\toprule
\textbf{Signal} & \textbf{Q2.5-14B} & \textbf{Q3-4B} \\
\midrule
$\Hhalf$ (ours) & .863 & \textbf{.864} \\
$\Hone$ (ours) & \textbf{.867} & \underline{.856} \\
$\Hord$ (ours) & \underline{.866} & .846 \\
Seq log-prob & .860 & .837 \\
Self-cons.$^\dag$ & .609 & .593 \\
Mean token entropy & .536 & .746 \\
Generation length & .808 & .833 \\
\bottomrule
\multicolumn{3}{@{}l}{\scriptsize $^\dag$Correlated SMC-particle agreement.}
\end{tabular}
\caption{Selective prediction AUROC on MATH500 with Qwen2.5-14B-Instruct and Qwen3-4B in non-thinking mode ($N{=}32$, $\alpha{=}4$, 500 problems).}
\label{tab:larger}
\end{table}

\section{Risk--Coverage Analysis}
\label{app:aurc}

Table~\ref{tab:aurc} reports the area under the risk--coverage curve (AURC), the error rate on the retained problems averaged over all coverage levels, for the settings of Table~\ref{tab:main}.
The best signal by AURC is the best signal by AUROC in all eight settings: $\Hhalf$ has the lowest AURC in six, an SRE signal is the best or second best in all eight, and $\Hhalf$ and $\Hone$ have a lower AURC than sequence log-probability in all eight.

\begin{table*}[t]
\centering
\small
\begin{tabular}{@{}lcccccccc@{}}
\toprule
 & \multicolumn{4}{c}{\textbf{MATH500}} & \multicolumn{4}{c}{\textbf{GSM8K}} \\
\cmidrule(lr){2-5} \cmidrule(lr){6-9}
\textbf{Signal} & \textbf{M-1.5B} & \textbf{M-7B} & \textbf{Q-7B} & \textbf{Llama} & \textbf{M-1.5B} & \textbf{M-7B} & \textbf{Q-7B} & \textbf{Llama} \\
\midrule
$\Hhalf$ (ours) & \textbf{11.9} & \textbf{8.0} & \textbf{10.2} & \textbf{23.3} & \textbf{7.7} & 5.2 & 9.1 & \textbf{7.4} \\
$\Hone$ (ours) & \underline{12.5} & \underline{9.0} & \underline{10.8} & \underline{24.6} & \underline{8.6} & \underline{5.0} & \underline{8.1} & \underline{7.8} \\
$\Hord$ (ours) & 12.9 & 9.5 & 11.6 & 24.8 & 8.9 & 5.6 & 10.3 & 8.6 \\
Seq log-prob & 13.0 & 9.3 & 12.2 & 25.0 & 8.8 & 5.9 & 10.5 & 9.0 \\
Self-cons.$^\dag$ & 23.6 & 16.3 & 21.6 & 47.3 & 11.9 & 8.4 & 9.6 & 11.0 \\
Mean token entropy & 20.1 & 15.2 & 29.0 & 63.0 & 11.4 & \textbf{4.2} & \textbf{7.9} & 8.0 \\
\bottomrule
\multicolumn{9}{@{}l}{\scriptsize $^\dag$Correlated SMC-particle agreement.}
\end{tabular}
\caption{Area under the risk--coverage curve ($\times 100$, lower is better; $N{=}32$, $\alpha{=}4$, 500 problems per setting). Ties are broken at random and averaged over 200 orderings. Best in \textbf{bold}, second best \underline{underlined}.}
\label{tab:aurc}
\end{table*}

\section{Estimator Variance Analysis}
\label{app:variance}

To characterize the reliability of the $\Hord$ estimator, we run each of 50 MATH500 problems $R{=}10$ times with different random seeds at $N{=}8$ and $N{=}32$ (Qwen2.5-Math-7B, $\alpha{=}4$).
Table~\ref{tab:variance} summarizes the results.

\begin{table}[h]
\centering
\small
\begin{tabular}{@{}ccccc@{}}
\toprule
$N$ & \textbf{Mean CV} & \textbf{Median CV} & \textbf{Split-half $\rho$} \\
\midrule
8 & 11.1\% & 8.3\% & 0.993 \\
32 & 7.8\% & 4.7\% & 0.993 \\
\bottomrule
\end{tabular}
\caption{Estimator variance of $\Hord$ across 10 repeated runs per problem (50 problems, Qwen2.5-Math-7B, $\alpha{=}4$).}
\label{tab:variance}
\end{table}

The coefficient of variation (CV) is small: the median CV is 8.3\% at $N{=}8$ and 4.7\% at $N{=}32$.
The \emph{ranking} of problems by $\Hord$ is almost perfectly stable: the Spearman rank correlation between the first and second halves of repeated runs is $\rho{=}0.993$ at both $N$ values, and two single runs agree at $\rho{=}0.964$ and $0.980$ on average.
This confirms that even a single $N{=}8$ probe reliably identifies which problems are hard.
On the 500 MATH500 problems the mean SRE from the standard estimator decreases with the particle budget (25.3, 23.2, and 22.6 at $N{=}8$, 32, and 64), consistent with the upward bias of Lemma~\ref{lem:smc_renyi}, which vanishes as $N$ grows; the stepwise accumulation follows the same trend (24.8, 23.3, and 22.9).

\section{Proof of Proposition~\ref{prop:chain}}
\label{app:chain}

Fix a prompt $x$ and a horizon $T$, with EOS treated as absorbing after termination, so that the next-token distribution after EOS is a point mass with zero entropy.
For $t = 0,\ldots,T$ let $Z_\beta^{(t)} = \sum_{y_{1:t}} \ptheta(y_{1:t}|x)^\beta$, so that $Z_\beta^{(0)} = 1$ and $Z_\beta^{(T)} = Z_\beta(x)$.
Factorizing $\ptheta(y_{1:t}|x) = \ptheta(y_{<t}|x)\,\ptheta(y_t|x,y_{<t})$ gives
\begin{align*}
Z_\beta^{(t)}
&= \sum_{y_{<t}} \ptheta(y_{<t}|x)^\beta \sum_{v} \ptheta(v|x,y_{<t})^\beta \\
&= Z_\beta^{(t-1)}\; \mathbb{E}_{y_{<t}\sim\pi_\beta^{(t-1)}}\!\left[\sum_{v} \ptheta(v|x,y_{<t})^\beta\right],
\end{align*}
and $\sum_{v} \ptheta(v|x,y_{<t})^\beta = e^{(1-\beta)h_\beta(y_{<t})}$ by the definition of $h_\beta$.
Taking logarithms, summing the telescoping terms over $t$, and dividing by $1-\beta$ gives the statement, since $H_\beta(\ptheta(\cdot|x)) = \frac{1}{1-\beta}\log Z_\beta^{(T)}$.
For the limit, $\pi_\beta^{(t-1)} \to \ptheta(y_{<t}|x)$ and $h_\beta \to H$ as $\beta \to 1$, and $\frac{1}{1-\beta}\log\mathbb{E}\big[e^{(1-\beta)h_\beta}\big] = \mathbb{E}[h_\beta] + O(1-\beta)$, so each increment converges to $\mathbb{E}_{y_{<t}\sim\ptheta}\left[H(Y_t|y_{<t},x)\right]$.

\section{Order Spectrum of the Stepwise Family}
\label{app:spectrum}

\begin{proposition}[Order spectrum]
\label{prop:spectrum}
Fix a per-problem run and a step $t$.
Draw a particle $I$ uniformly from $A_t$ and a token $V\sim\ptheta(\cdot|x,y^{(I)}_{<t})$, and let $X_t=\ptheta(V|x,y^{(I)}_{<t})$ be the probability the model gives to its own next token.
Let $M_r(X)=(\mathbb{E}[X^r])^{1/r}$ be the power mean of order $r$, with $M_0(X)=\exp\mathbb{E}[\log X]$, and let $h_\infty=-\log\max_v\ptheta(v|\cdot)$.
\begin{enumerate}[leftmargin=1.7em]
\item[(a)] The $t$-th increment of Eq.~\ref{eq:family} equals $-\log M_{\beta-1}(X_t)$, so $SRE_\beta=\sum_t-\log M_{\beta-1}(X_t)$.
\item[(b)] On every per-problem run $SRE_\beta$ is nonincreasing in $\beta$, with $SRE_\beta\to T\log|\mathcal{V}|$ as $\beta\to0$ and $SRE_\beta\to\sum_t\min_{i\in A_t}h_\infty(y^{(i)}_{<t})$ as $\beta\to\infty$.
\item[(c)] For $\beta>1$ every next-token distribution satisfies $h_\infty\le h_\beta\le\frac{\beta}{\beta-1}h_\infty$.
Hence each increment of Eq.~\ref{eq:family} is at least $\min_{i\in A_t} h_\infty(y^{(i)}_{<t})$ and at most $\frac{\beta}{\beta-1}$ times the mean of $h_\infty(y^{(i)}_{<t})$ over $A_t$, and at the sequence level $H_\infty(\ptheta)\le H_\alpha(\ptheta)\le\frac{\alpha}{\alpha-1}H_\infty(\ptheta)$ with $H_\infty=-\log\max_y\ptheta(y|x)$.
\item[(d)] For $\beta<1$ and a next-token distribution with top probability $1-\varepsilon$,
$(1-\varepsilon)^\beta+\varepsilon^\beta \le e^{(1-\beta)h_\beta} \le (1-\varepsilon)^\beta+\varepsilon^\beta(|\mathcal{V}|-1)^{1-\beta}$.
The upper end is attained when the remaining probability mass is uniform and the lower end, for $\varepsilon\le1/2$, when it sits on one token.
For $\beta=1$ the range of $h_1$ is $[h_b(\varepsilon),\,h_b(\varepsilon)+\varepsilon\log(|\mathcal{V}|-1)]$ with $h_b$ the binary entropy.
\end{enumerate}
\end{proposition}

\begin{proof}
(a) Write $p_i=\ptheta(\cdot|x,y^{(i)}_{<t})$. Then
$\mathbb{E}[X_t^{\beta-1}]=\frac{1}{|A_t|}\sum_{i\in A_t}\sum_v p_i(v)\,p_i(v)^{\beta-1}=\frac{1}{|A_t|}\sum_{i\in A_t}e^{(1-\beta)h_\beta(y^{(i)}_{<t})}$,
so the $t$-th increment of Eq.~\ref{eq:family} is $\frac{1}{1-\beta}\log\mathbb{E}[X_t^{\beta-1}]=-\log M_{\beta-1}(X_t)$.
For $\beta=1$ the increment is $\frac{1}{|A_t|}\sum_i H(p_i)=\mathbb{E}[-\log X_t]=-\log M_0(X_t)$.

(b) The power mean $M_r(X)$ of a positive random variable is nondecreasing in $r$, so each increment is nonincreasing in $\beta$ and so is the sum.
As $r\to\infty$, $M_r(X_t)\to\max_{i,v}p_i(v)$, which gives $\min_i h_\infty(y^{(i)}_{<t})$.
As $\beta\to0$ the increment tends to $\log\frac{1}{|A_t|}\sum_i|\mathrm{supp}\,p_i|=\log|\mathcal{V}|$ for a full-support softmax.

(c) Let $m=\max_v p(v)$. Then $m^\beta\le\sum_v p(v)^\beta\le m^{\beta-1}\sum_v p(v)=m^{\beta-1}$.
Applying $\frac{1}{1-\beta}\log$, which reverses inequalities for $\beta>1$, gives $-\log m\le h_\beta\le\frac{\beta}{\beta-1}(-\log m)$.
For the particle aggregate, write $h_i=h_\beta(y^{(i)}_{<t})$ and $\bar h$ for their mean over $A_t$. Convexity of $x\mapsto e^{(1-\beta)x}$ gives $\frac{1}{|A_t|}\sum_i e^{(1-\beta)h_i}\ge e^{(1-\beta)\bar h}$, so the increment is at most $\bar h$, which is at most $\frac{\beta}{\beta-1}$ times the mean of $h_\infty$.
Also $\frac{1}{|A_t|}\sum_i e^{(1-\beta)h_i}\le e^{(1-\beta)\min_i h_i}$, so the increment is at least $\min_i h_i\ge\min_i h_\infty(y^{(i)}_{<t})$.
The sequence-level statement is the first display applied to $\ptheta(\cdot|x)$ over completions.

(d) Let the other probabilities be $q_1,\dots,q_K$ with $K=|\mathcal{V}|-1$ and $\sum_j q_j=\varepsilon$.
For $0<\beta<1$, $x\mapsto x^\beta$ is concave with value 0 at 0 and hence subadditive, so $\sum_j q_j^\beta\ge\varepsilon^\beta$; by Jensen's inequality $\sum_j q_j^\beta\le K(\varepsilon/K)^\beta=\varepsilon^\beta K^{1-\beta}$.
Adding $(1-\varepsilon)^\beta$ gives the two bounds.
The $\beta=1$ range follows from $H(p)=h_b(\varepsilon)+\varepsilon H(q/\varepsilon)$ and $0\le H(q/\varepsilon)\le\log K$.
\end{proof}

The logged per-problem runs agree with the proposition.
Part (b) and the bounds of part (c) hold in every one of the 5{,}500 per-problem runs that log the order spectrum.
The order-4 increments, averaged over particles, sum to 1.25 times the corresponding top-token terms (median per setting between 1.249 and 1.257), inside the range $[1, 4/3]$.
The Spearman correlation of $SRE_\beta$ with the number of decoding steps rises to at least 0.93 at order $1/4$, the direction of the $\beta\to0$ limit.

\section{Proof of Proposition~\ref{prop:search}}
\label{app:search}

Conditioned on $Y{=}y$, the number of independent draws from $q$ until $y$ first appears is geometric with mean $1/q(y)$, so the expected number of draws is $\sum_y p(y)/q(y)$, which is infinite when $q(y){=}0$ for some $y$ with $p(y)>0$.
By the Cauchy--Schwarz inequality,
\begin{align*}
e^{H_{1/2}(p)}
&= \Big(\sum_y p(y)^{1/2}\Big)^{2} \\
&= \Big(\sum_y \sqrt{\tfrac{p(y)}{q(y)}}\,\sqrt{q(y)}\Big)^{2} \\
&\leq \sum_y \frac{p(y)}{q(y)} \sum_y q(y)
 = \sum_y \frac{p(y)}{q(y)},
\end{align*}
with equality if and only if $p(y)/q(y) \propto q(y)$, that is, $q \propto p^{1/2}$.
\citet{arikan1996inequality} shows that the expected number of guesses when candidates are tried in order of decreasing probability lies between $e^{H_{1/2}(p)}/(1+\ln M)$ and $e^{H_{1/2}(p)}$ for $M$ outcomes, so the same quantity governs guesswork.

\begin{proposition}[Guesswork of the power distribution]
\label{prop:guess}
For $\alpha>1$, $H_\alpha(\ptheta(\cdot|x))=H_{1/\alpha}(\pi_\alpha(\cdot|x))$.
Consequently, if $Y\sim\pi_\alpha$ takes $M$ values and $G^\star(Y)$ is the number of guesses needed when completions are tried in order of decreasing probability,
\[
\frac{e^{H_\alpha(\ptheta)}}{1+\ln M}\le\big(\mathbb{E}[G^\star(Y)^{\alpha-1}]\big)^{\frac{1}{\alpha-1}}\le e^{H_\alpha(\ptheta)}.
\]
\end{proposition}

\begin{proof}
$\sum_y\pi_\alpha(y)^{1/\alpha}=\sum_y\ptheta(y|x)/Z_\alpha^{1/\alpha}=Z_\alpha^{-1/\alpha}$, so
$H_{1/\alpha}(\pi_\alpha)=\frac{1}{1-1/\alpha}\log Z_\alpha^{-1/\alpha}=\frac{1}{1-\alpha}\log Z_\alpha=H_\alpha(\ptheta)$.
The inequality of \citet{arikan1996inequality}, for any $\rho>0$ and any distribution $P$ on $M$ outcomes, reads
$(1+\ln M)^{-\rho}e^{\rho H_{1/(1+\rho)}(P)}\le\mathbb{E}[G^\star(X)^\rho]\le e^{\rho H_{1/(1+\rho)}(P)}$.
Take $P=\pi_\alpha$ and $\rho=\alpha-1$, so that $1/(1+\rho)=1/\alpha$, substitute the identity, and take the power $1/\rho$.
\end{proof}

\section{Proof of Lemma~\ref{lem:smc_renyi}}
\label{app:proof}

Let $\gamma_t(y_{1:t})$, $t=0,\dots,T$, be the positive unnormalized intermediate targets of the sampler, with $\gamma_0=1$ and $\gamma_T(y)=\ptheta(y|x)^\alpha$, let $q_t(\cdot|y_{<t})$ be the proposals, and let $\omega_t=\gamma_t(y_{1:t})/[\gamma_{t-1}(y_{<t})\,q_t(y_t|y_{<t})]$ be the incremental weights.
Resampling is unbiased: the expected number of copies of particle $j$ is $N$ times its normalized weight.
Power-SMC is the case $\gamma_t=\ptheta(y_{1:t}|x)^{\alpha_t}$ with the exponent ramp $\alpha_t$ of Appendix~\ref{app:resampling}, a nucleus-truncated proposal, and systematic resampling at block boundaries when the effective sample size is below $N/2$.

\begin{proof}
Write the particle system after step $t$ as $(y^{(i)}_{1:t},w^{(i)}_t)_{i\le N}$, where $w$ is the unnormalized weight accumulated since the last resampling event, and let $C_t$ be the product of the averages $\frac{1}{N}\sum_k w^{(k)}$ taken at the resampling events so far ($C_0=1$, $w_0^{(i)}=1$).
By the telescoping of Appendix~\ref{app:resampling}, $\widehat{Z}_{\alpha,N}=C_T\frac{1}{N}\sum_i w_T^{(i)}$.
For a function $\varphi$ of length-$t$ prefixes put
\begin{align*}
\Phi_t(\varphi)&=C_t\,\tfrac{1}{N}\textstyle\sum_i w_t^{(i)}\varphi(y_{1:t}^{(i)}),\\
\gamma_t(\varphi)&=\textstyle\sum_{y_{1:t}\in\mathcal{S}_t}\gamma_t(y_{1:t})\,\varphi(y_{1:t}),
\end{align*}
where $\mathcal{S}_t$ is the set of prefixes that the proposals can produce.
We show $\mathbb{E}[\Phi_t(\varphi)]=\gamma_t(\varphi)$ for all $t$ and $\varphi$ by induction; for $t=0$ both sides equal $\varphi(\emptyset)$.

\emph{Propagation.} Given the system after step $t-1$, whose history we denote $\mathcal{F}_{t-1}$, particle $i$ draws $y_t\sim q_t(\cdot|y^{(i)}_{<t})$ and $w_t^{(i)}=w_{t-1}^{(i)}\omega_t$, so
$\mathbb{E}[w_t^{(i)}\varphi(y^{(i)}_{1:t})\mid\mathcal{F}_{t-1}]=w^{(i)}_{t-1}\tilde\varphi(y^{(i)}_{<t})$ with
$\tilde\varphi(y_{<t})=\sum_{v:\,q_t(v|y_{<t})>0}\frac{\gamma_t(y_{<t},v)}{\gamma_{t-1}(y_{<t})}\varphi(y_{<t},v)$.
Hence $\mathbb{E}[\Phi_t(\varphi)\mid\mathcal{F}_{t-1}]=\Phi_{t-1}(\tilde\varphi)$, whose expectation is $\gamma_{t-1}(\tilde\varphi)=\gamma_t(\varphi)$ by the induction hypothesis.
A finished particle is the special case of a deterministic extension, and the final exponent correction of the ramp is a reweighting by $\gamma'(y)/\gamma(y)$ with no new token, for which $\mathbb{E}[\Phi'(\varphi)]=\gamma(\varphi\gamma'/\gamma)=\gamma'(\varphi)$.

\emph{Resampling.} The decision to resample is a function of the system at time $t$.
If it is taken, the new particles are $\hat y^{(i)}=y^{(A_i)}$ with unit weights and $\hat C=C_t\frac{1}{N}\sum_k w^{(k)}_t$.
Since the expected number of copies of particle $j$ is $N w^{(j)}_t/\sum_k w^{(k)}_t$,
$\mathbb{E}[\hat C\,\frac{1}{N}\sum_i\varphi(\hat y^{(i)})\mid\mathcal{F}_t]=\Phi_t(\varphi)$, so the claim is preserved.

\emph{Item 1.} With $\varphi\equiv1$ at $t=T$, $\mathbb{E}[\widehat{Z}_{\alpha,N}]=\gamma_T(1)=\sum_{y\in\mathcal{S}_T}\ptheta(y|x)^\alpha\le Z_\alpha$, with equality when the proposals have full support.

\emph{Item 2.} By Jensen's inequality $\mathbb{E}[\log\widehat{Z}_{\alpha,N}]\le\log\mathbb{E}[\widehat{Z}_{\alpha,N}]\le\log Z_\alpha$; multiplying by $1/(1-\alpha)<0$ reverses the inequality.

\emph{Item 3.} By Markov's inequality $\Pr[\widehat{Z}_{\alpha,N}\ge Z_\alpha/\delta]\le\delta\,\mathbb{E}[\widehat{Z}_{\alpha,N}]/Z_\alpha\le\delta$.
On the complement $\log\widehat{Z}_{\alpha,N}<\log Z_\alpha+\log(1/\delta)$, and dividing by $1-\alpha<0$ gives $\widehat{\mathrm{SRE}}_{\alpha,N}>H_\alpha(\ptheta(\cdot|x))-\log(1/\delta)/(\alpha-1)$.

\emph{Consistency.} With full-support proposals and multinomial resampling at fixed times, $\widehat{Z}_{\alpha,N}\to Z_\alpha$ in probability for a fixed horizon \citep{delmoral2004feynman,chopin2020introduction}, and the continuous mapping theorem applied to $\frac{1}{1-\alpha}\log$ gives the limit of $\widehat{\mathrm{SRE}}_{\alpha,N}$.
\end{proof}

\section{Generality Beyond Power-SMC}
\label{app:generality}

The estimator requires only the incremental weights $w_t^{(i)}$ of a sampler whose final target is the power distribution $\pi_\alpha$ and whose resampling is unbiased.
For any such sampler, Lemma~\ref{lem:smc_renyi} applies to the estimate of $Z_\alpha$ from the weights, from which $\Halpha = \log\Zalpha/(1-\alpha)$.

\paragraph{APPS.}
APPS \citep{nguyen2026apps} is a blockwise auxiliary-particle sampler.
Its propagation step assigns each particle the proposal-corrected power weight
\begin{align*}
\log w^{(i)} \mathrel{+}= {}& \alpha \log \ptheta\big(b_j^{(i)} \mid x_{\mathrm{prev}}^{(i)}\big) \\
& - \log q_j\big(b_j^{(i)} \mid x_{\mathrm{prev}}^{(i)}\big)
\end{align*}
for block $b_j^{(i)}$ under proposal $q_j$.
These are the incremental weights of an SMC sampler whose final target is $\ptheta^\alpha$, so the estimator applies with no extra model calls.
Lemma~\ref{lem:smc_renyi} holds for arbitrary positive intermediate targets, so the future-value potential that APPS uses at resampling boundaries changes the intermediate targets only.
The estimator accumulates, at each resampling step, the average of the weights that the sampler resamples with, and the potentials cancel in the product because the final potential is 1.
The estimand is $Z_\alpha$.

\paragraph{Rollout-based power sampling.}
Scalable Power Sampling \citep{ji2026scalable} samples autoregressively from $p^{(\mathrm{pow})}_\alpha(x_t \mid \cdot) \propto \ptheta(x_t \mid \cdot)^\alpha\, \zeta_t(x_t)$ with a future-value factor $\zeta_t$ estimated by rollouts.
Each per-token normalizer is $\Gamma_t(y_{<t})=\sum_v\ptheta(v|y_{<t})^\alpha\zeta_t(y_{<t},v)$.
With the exact future value $\zeta_t(y_{\le t})=\sum_{y_{>t}}\ptheta(y_{>t}|y_{\le t})^\alpha$ the first normalizer is the sequence-level normalizing constant, $\Gamma_1=\sum_v\ptheta(v)^\alpha\sum_{y_{>1}}\ptheta(y_{>1}|v)^\alpha=Z_\alpha$, so $\Halpha=\frac{1}{1-\alpha}\log\widehat\Gamma_1$ is available after the rollouts of the first step.
With estimated future values the sampler draws $y$ with probability $q(y)=\prod_t\ptheta(y_t|y_{<t})^\alpha\zeta_t(y_{\le t})/\Gamma_t(y_{<t})$, and
$\mathbb{E}_{y\sim q}\big[\prod_{t}\Gamma_t(y_{<t})/\zeta_t(y_{\le t})\big]=\sum_y\prod_t\ptheta(y_t|y_{<t})^\alpha=Z_\alpha$,
so the mean of this weight over the sampled completions is an unbiased estimate of $Z_\alpha$ from quantities the method already computes.

\section{Sensitivity to the Sharpening Exponent}
\label{app:alpha}

\begin{figure}[h]
    \centering
\includegraphics[width=0.92\linewidth]{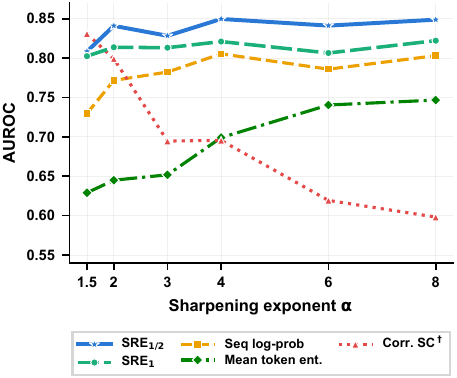}
    \caption{Selective prediction AUROC vs.\ $\alpha$ (MATH500, Qwen2.5-Math-7B, $N{=}32$). $^\dagger$SC is particle agreement (correlated self-consistency).}
    \label{fig:alpha_ablation}
\end{figure}

\section{Length Control}
\label{app:length}

\begin{table}[h]
\centering
\small
\begin{tabular}{@{}lcccc@{}}
\toprule
 & \multicolumn{2}{c}{\textbf{MATH500}} & \multicolumn{2}{c}{\textbf{GSM8K}} \\
\cmidrule(lr){2-3} \cmidrule(lr){4-5}
\textbf{Length quartile} & \textbf{Low} & \textbf{High} & \textbf{Low} & \textbf{High} \\
\midrule
Q1 (shortest) & 96.8 & 69.4 & 95.2 & 66.1 \\
Q2 & 95.2 & 77.4 & 100.0 & 82.3 \\
Q3 & 92.1 & 67.7 & 95.2 & 82.3 \\
Q4 (longest) & 73.0 & 33.9 & 92.1 & 75.8 \\
\bottomrule
\end{tabular}
\caption{Accuracy (\%) of the below-median (Low) and above-median (High) $\Hord$ halves within each quartile of generation length (Qwen2.5-Math-7B, $N{=}32$, $\alpha{=}4$).}
\label{tab:length_strat}
\end{table}

Table~\ref{tab:length_strat_all} applies the same length control to every signal: the AUROC is computed over correct--incorrect pairs inside the same quartile of generation length.
$\Hhalf$ is above sequence log-probability inside length quartiles in all eight settings, by 0.020 to 0.052, and $\Hone$ in seven.
$\Hhalf$ is usually less driven by length than log-probability is: its Spearman correlation with the length of the returned completion is lower than that of log-probability in six of the eight settings.
Removing the problems whose decoding reaches the 2{,}048-token limit (82 of 500 for Llama-3.1-8B on MATH500, at most 27 elsewhere) leaves $\Hhalf$ above sequence log-probability in all eight settings, by 0.015 to 0.048.

\begin{table*}[t]
\centering
\small
\begin{tabular}{@{}lcccccccc@{}}
\toprule
 & \multicolumn{4}{c}{\textbf{MATH500}} & \multicolumn{4}{c}{\textbf{GSM8K}} \\
\cmidrule(lr){2-5} \cmidrule(lr){6-9}
\textbf{Signal} & \textbf{M-1.5B} & \textbf{M-7B} & \textbf{Q-7B} & \textbf{Llama} & \textbf{M-1.5B} & \textbf{M-7B} & \textbf{Q-7B} & \textbf{Llama} \\
\midrule
$\Hhalf$ (ours) & .821 & .823 & .804 & .741 & .781 & .795 & .701 & .742 \\
$\Hone$ (ours) & .813 & .789 & .789 & .708 & .757 & .811 & .731 & .721 \\
$\Hord$ (ours) & .799 & .773 & .770 & .713 & .738 & .790 & .677 & .711 \\
Seq log-prob & .791 & .773 & .760 & .709 & .729 & .776 & .677 & .708 \\
Mean token entropy & .755 & .747 & .715 & .649 & .723 & .814 & .719 & .690 \\
\bottomrule
\end{tabular}
\caption{Length-stratified selective prediction AUROC ($N{=}32$, $\alpha{=}4$, 500 problems per setting): correct--incorrect pairs are compared only inside the same quartile of generation length.}
\label{tab:length_strat_all}
\end{table*}

\section{Full Adaptive Particle Allocation Sweep}
\label{app:sweep}

\begin{table}[h]
\centering
\small
\setlength{\tabcolsep}{3pt}
\begin{tabular}{@{}lccccc@{}}
\toprule
 & \multicolumn{4}{c}{\textbf{Math-7B}} & \textbf{M-1.5B} \\
\cmidrule(lr){2-5} \cmidrule(lr){6-6}
\textbf{Gating signal} & \textbf{s42} & \textbf{s43} & \textbf{s44} & \textbf{mean} & \textbf{s42} \\
\midrule
$\Hhalf$ (ours) & 77.01 & 76.26 & 75.68 & 76.31 & 69.39 \\
$\Hord$ (ours) & 76.88 & 76.10 & 75.81 & 76.26 & 69.19 \\
$\Hone$ (ours) & 76.84 & 76.17 & 75.76 & 76.26 & 69.27 \\
Seq log-prob & 76.76 & 76.11 & 75.75 & 76.21 & 69.25 \\
Answer entropy & 76.69 & 76.26 & 76.75 & 76.57 & 69.10 \\
Self-Certainty & 76.50 & 76.86 & 76.09 & 76.48 & 70.06 \\
Mean token entropy & 76.29 & 76.06 & 76.23 & 76.19 & 69.60 \\
\midrule
Random escalation & 76.00 & 76.10 & 75.50 & 75.87 & 68.90 \\
\bottomrule
\end{tabular}
\caption{Gating signals for adaptive particle allocation, all computed from the same $N{=}8$ probe (MATH500, 500 problems; s42 to s44: seeds). Entries are the accuracy (\%) of the two-pass policy averaged over escalation rates from 0\% to 100\% in steps of 5\%.}
\label{tab:gating}
\end{table}

\begin{table}[h]
\centering
\small
\begin{tabular}{@{}lccc@{}}
\toprule
\textbf{Method} & \textbf{Acc (\%)} & \textbf{Compute} & \textbf{Savings} \\
\midrule
Uniform $N{=}8$ & 74.8 & $1.0\times$ & --- \\
Uniform $N{=}32$ & 75.8 & $4.0\times$ & --- \\
\midrule
Adaptive (Esc.\ 30\%) & 76.4 & $3.4\times$ & 57.5\% \\
Adaptive (Esc.\ 40\%) & 77.2 & $4.2\times$ & 47.5\% \\
Adaptive (Esc.\ 45\%) & \textbf{77.8} & $4.6\times$ & 42.5\% \\
Adaptive (Esc.\ 70\%) & \textbf{77.8} & $6.6\times$ & 17.5\% \\
\midrule
Uniform $N{=}64$ & 77.2 & $8.0\times$ & 0\% \\
\bottomrule
\end{tabular}
\caption{Adaptive particle allocation vs.\ uniform baselines on MATH500 (500 problems, Qwen2.5-Math-7B). Esc.\ $X$\%: escalate the
$X$\% of problems with the highest $\Hord$ to $N{=}64$.
Compute is in $N{=}8$ units and adaptive rows include the probe cost. Savings are relative to uniform $N{=}64$.}
\label{tab:pareto}
\end{table}

Table~\ref{tab:seeds} repeats the experiment with three seeds.
Over the three seeds, the mid and high terciles gain more than the low tercile in 99.4\% of 10,000 bootstrap replicates of the problems.
On Qwen2.5-Math-1.5B ($N{=}8$: 67.2\%, $N{=}64$: 70.6\%) the tercile gains are $+$0.6, $+$3.6, and $+$6.0.
The savings statistic of Table~\ref{tab:seeds} is 34.2\% for $\Hord$ and $\Hone$, 32.5\% for the standard estimator, sequence log-probability and Self-Certainty, 22.5\% for $\Hhalf$, 20.8\% for answer entropy, 7.5\% for mean token entropy, and $-$0.8\% for random escalation (mean of 5{,}000 random orders); $\Hord$, $\Hone$, the standard estimator and log-probability are the only gating signals with a positive saving on all three seeds.
With one escalation rate for all seeds, escalating the 60\% highest-$\Hord$ problems gives the three-seed mean accuracy of uniform $N{=}64$ (76.6\%) with 27.5\% fewer particles, and 50\% gives 76.5\% with 37.5\% fewer.

\begin{table}[h]
\centering
\small
\setlength{\tabcolsep}{4pt}
\begin{tabular}{@{}lcccccc@{}}
\toprule
 & \multicolumn{2}{c}{\textbf{Accuracy}} & \multicolumn{3}{c}{\textbf{Gain by tercile}} & \\
\cmidrule(lr){2-3} \cmidrule(lr){4-6}
\textbf{Seed} & $N{=}8$ & $N{=}64$ & Low & Mid & High & \textbf{Savings} \\
\midrule
42 & 74.8 & 77.2 & $-$1.2 & $+$3.0 & $+$5.4 & 47.5\% \\
43 & 75.8 & 76.4 & $-$0.6 & $+$3.0 & $-$0.6 & 22.5\% \\
44 & 74.8 & 76.2 & $+$0.0 & $+$1.2 & $+$3.0 & 32.5\% \\
\midrule
Mean & 75.1 & 76.6 & $-$0.6 & $+$2.4 & $+$2.6 & 34.2\% \\
\bottomrule
\end{tabular}
\caption{Adaptive particle allocation with three seeds (MATH500, 500 problems, Qwen2.5-Math-7B, $\alpha{=}4$). Gain: accuracy at $N{=}64$ minus accuracy at $N{=}8$ within each tercile of probe $\Hord$. Savings: particles saved relative to uniform $N{=}64$ at the smallest escalation rate from which the two-pass policy matches uniform $N{=}64$ accuracy.}
\label{tab:seeds}
\end{table}

Table~\ref{tab:sweep} reports the full threshold sweep behind Table~\ref{tab:pareto}.
Every escalation rate from 40\% to 100\% matches or exceeds uniform $N{=}64$ accuracy (77.2\%).

\begin{table}[h]
\centering
\small
\begin{tabular}{@{}rcc@{\hspace{1.5em}}rcc@{}}
\toprule
\textbf{Esc.} & \textbf{Acc} & \textbf{Comp.} & \textbf{Esc.} & \textbf{Acc} & \textbf{Comp.} \\
\midrule
0\% & 74.8 & $1.0\times$ & 55\% & 77.6 & $5.4\times$ \\
5\% & 74.6 & $1.4\times$ & 60\% & 77.4 & $5.8\times$ \\
10\% & 75.2 & $1.8\times$ & 65\% & 77.6 & $6.2\times$ \\
15\% & 75.4 & $2.2\times$ & 70\% & 77.8 & $6.6\times$ \\
20\% & 76.4 & $2.6\times$ & 75\% & 77.6 & $7.0\times$ \\
25\% & 76.6 & $3.0\times$ & 80\% & 77.6 & $7.4\times$ \\
30\% & 76.4 & $3.4\times$ & 85\% & 77.6 & $7.8\times$ \\
35\% & 77.0 & $3.8\times$ & 90\% & 77.4 & $8.2\times$ \\
40\% & 77.2 & $4.2\times$ & 95\% & 77.4 & $8.6\times$ \\
45\% & 77.8 & $4.6\times$ & 100\% & 77.2 & $9.0\times$ \\
50\% & 77.8 & $5.0\times$ & & & \\
\bottomrule
\end{tabular}
\caption{Accuracy (\%) and average compute (in $N{=}8$ units, probe included) as a function of the fraction of highest-$\Hord$ problems escalated to $N{=}64$ (MATH500, 500 problems, Qwen2.5-Math-7B). Uniform $N{=}64$: 77.2\% at $8.0\times$.}
\label{tab:sweep}
\end{table}

\section{Three-Way Policy: Answer, Escalate, or Abstain}
\label{app:threeway}

The probe signal can drive a three-way decision.
We rank problems by probe $\Hord$, abstain on the most uncertain ones, escalate the next $e$\% of all problems to $N{=}64$, and keep the probe answer for the rest.
Table~\ref{tab:threeway} reports the accuracy on the answered problems, averaged over three seeds, against uniform $N{=}64$, which abstains by its own $\Hord$.
At 80\% coverage the three-way policy with 40\% escalation matches uniform $N{=}64$ with 47.5\% fewer particles, and at 60\% coverage 20\% escalation suffices.

\begin{table}[h]
\centering
\small
\begin{tabular}{@{}lccc@{}}
\toprule
 & & \multicolumn{2}{c}{\textbf{Accuracy at coverage}} \\
\cmidrule(lr){3-4}
\textbf{Policy} & \textbf{Compute} & \textbf{80\%} & \textbf{60\%} \\
\midrule
Probe only ($N{=}8$) & $1.0\times$ & 84.2 & 89.2 \\
Three-way, $e{=}20$ & $2.6\times$ & 84.5 & 90.0 \\
Three-way, $e{=}40$ & $4.2\times$ & 85.1 & 90.3 \\
\midrule
Uniform $N{=}64$ & $8.0\times$ & 85.0 & 89.9 \\
\bottomrule
\end{tabular}
\caption{Three-way policy on MATH500 (500 problems, Qwen2.5-Math-7B, $\alpha{=}4$, mean over three seeds). Accuracy (\%) on the answered problems when the policy abstains on the most uncertain 20\% or 40\%. Compute is in $N{=}8$ units and includes the probe.}
\label{tab:threeway}
\end{table}

\end{document}